\documentclass[twocolumn]{article}

\usepackage{changepage}
\usepackage{arxiv}
\usepackage[T1]{fontenc}
\usepackage[utf8]{inputenc}
\usepackage{graphicx}
\usepackage{amssymb, amsthm, amsmath} 
\usepackage{algorithm}
\usepackage{algpseudocode}
\usepackage{url}
\usepackage{makecell}
\usepackage{booktabs}
\usepackage{multirow}
\usepackage[hidelinks]{hyperref}
\usepackage{microtype}
\usepackage[table]{xcolor}

\newtheorem{theorem}{Theorem}

\theoremstyle{remark}

\title{CIFNet: An Analytic Neural Learning Framework for Efficient and Calibrated Class-Incremental Learning}

\author{
  \begin{tabular}{c c}
    \href{https://orcid.org/0009-0005-0592-5760}{\includegraphics[scale=0.06]{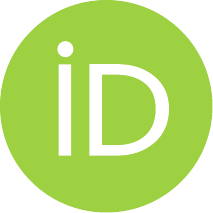}\hspace{1mm}Alejandro Dopico-Castro}\thanks{Corresponding author: \texttt{alejandro.dopico2@udc.es}} &
    \href{https://orcid.org/0000-0003-4203-8720}{\includegraphics[scale=0.06]{orcid.pdf}\hspace{1mm}Oscar Fontenla-Romero} \\[0.5em]
    \href{https://orcid.org/0000-0001-8901-5441}{\includegraphics[scale=0.06]{orcid.pdf}\hspace{1mm}Bertha Guijarro-Berdiñas} &
    \href{https://orcid.org/0000-0003-0950-0012}{\includegraphics[scale=0.06]{orcid.pdf}\hspace{1mm}Amparo Alonso-Betanzos}
  \end{tabular} \\[1em]
  Universidade da Coruña, CITIC \\
  Facultade de Informática, Campus de Elviña s/n, 15071 A Coruña, Spain \\
  \texttt{\{alejandro.dopico2, oscar.fontenla, berta.guijarro, amparo.alonso.betanzos\}@udc.es}
}

\date{}

\renewcommand{\shorttitle}{CIFNet: Analytic Framework for Class-Incremental Learning}

\hypersetup{
    pdftitle={CIFNet: An Analytic Neural Learning Framework for Efficient and Calibrated Class-Incremental Learning},
    pdfsubject={cs.LG, cs.AI},
    pdfauthor={Alejandro Dopico-Castro et al.},
    pdfkeywords={Class-Incremental Learning, Analytic Learning, Regularised Recursive Least-Squares, Neuron Calibration, Energy Efficiency},
}

\begin{document}

\twocolumn[
    \maketitle
    \begin{abstract}

    Class-Incremental Learning (CIL) in deep neural networks is conventionally framed as an iterative gradient-based optimization problem, incurring high computational cost, hyperparameter sensitivity, and risk of catastrophic forgetting. In this work, we demonstrate that when leveraging frozen pre-trained representations, CIL can be solved as a sequence of deterministic, closed-form classifier adaptations without backpropagation or iterative convergence. We propose CIFNet, an analytic neural learning framework built upon Regularised Recursive Least-Squares (RRLS). CIFNet updates classifier weights via an exact, closed-form ridge-regression solution operating in a stationary embedding space. To counteract the structural initialisation bias that arises when newly expanded output neurons are introduced without exposure to past-class evidence, CIFNet incorporates a lightweight calibration buffer in latent space alongside density-aware oversampling, ensuring globally balanced decision boundaries without raw image storage or gradient updates. Extensive evaluations across CIFAR-100, ImageNet-100, and CORe50 show that CIFNet achieves predictive accuracy competitive with iterative CIL baselines while maintaining strictly monotonic, smooth learning trajectories free from intermediate performance collapse. Furthermore, by replacing epoch-wise backpropagation with closed-form moment accumulation, CIFNet achieves up to 20$\times$ reduction in energy consumption. These findings establish calibrated analytic learning as an efficient, stable, and mathematically grounded paradigm for continual adaptation in neural networks.

    \vspace{2em}
    \end{abstract}
]

\section{Introduction}

Class Incremental Learning (CIL) \cite{schlimmer1986case, Rebuffi2017} represents a critical capability for modern intelligent systems: a model must learn to recognise an expanding set of classes as they arrive sequentially, without forgetting those learned earlier. The dominant paradigm in CIL addresses this by fine-tuning the entire network through iterative backpropagation, employing regularisation, replay, or architectural expansion to protect previously acquired knowledge \cite{zhou2024cil_survey}.

However, this approach carries a severe, hidden cost. At every incremental step, the backbone, despite already encoding high-quality transferable features, is subjected to repeated gradient updates. This consumes energy proportional to the number of epochs, tasks, and parameters involved. Over a long incremental horizon, this cost compounds into an environmental and computational burden that is rarely quantified but is far from negligible \cite{strubell2019energy}.

This burden is particularly concerning given the current landscape of vision-based systems. Large-scale pre-trained visual models, such as ResNets or Vision Transformers trained on ImageNet, encode representations that transfer reliably across dozens of downstream tasks, often matching or exceeding the performance of models trained from scratch with far more data \cite{zhou2025revisiting}. This representational richness is the product of an enormous initial computational investment, an investment that has already been made and should not need to be repeated for every new downstream application. 

More fundamentally, continuous fine-tuning of a strong, pre-trained backbone may be structurally unnecessary. If the representation is already expressive enough to separate the incremental classes in feature space, the hard problem of visual representation learning has already been solved. What remains is a comparatively simpler system-level challenge: updating a linear classifier to reflect the growing label space. This observation motivates the approach proposed in this work, which exploits the PTM rather than modifying it, treating incremental learning as a sequence of efficient classifier adaptations rather than a sequence of expensive re-training episodes. From a neural network learning perspective, this reframing has a precise consequence: catastrophic forgetting, which arises from the distributed nature of weight-based memory in gradient-trained networks, can be structurally avoided at the classifier level when the underlying representation is held fixed. The stability-plasticity dilemma then reduces from a problem of network-wide optimisation to one of incremental linear estimation on a stationary feature space, a fundamentally simpler and more tractable problem than the general case addressed by iterative continual learning methods.

This work introduces \textbf{CIFNet} (\textbf{C}lass \textbf{I}ncremental \textbf{F}rugal \textbf{Net}work), a CIL framework built on three core principles. First, a pre-trained backbone is fixed throughout incremental deployment: it serves as a stationary feature extractor, avoiding representation drift and reducing feature computation to a single forward pass per sample. Second, the classifier is updated analytically via a Regularised Recursive Least-Squares (RRLS) formulation that yields the globally optimal weight matrix in closed form, completely removing Stochastic Gradient Descent (SGD) backpropagation, learning rate schedules, or convergence criteria. Third, a compact calibration buffer of stored embeddings corrects the structural initialisation bias that arises when new output neurons are introduced without exposure to past-class evidence, a systemic problem distinct from catastrophic forgetting.

CIFNet is most closely related to recent analytic CIL methods \cite{zhuang2022acil, zhuang2024ds}, which share the frozen-backbone, closed-form classifier premise and demonstrate that analytic learning can match or exceed the accuracy of iterative approaches in long-horizon settings. Within this family, CIFNet adopts a distinct methodology: rather than artificially augmenting the embedding space via random projections, it operates directly on native backbone embeddings and introduces a lightweight calibration mechanism to address the class-expansion initialisation problem. Moreover, to the best of the authors' knowledge, CIFNet additionally provides the first systematic energy and sustainability analysis of this methodological family.

Sustainability is a primary motivation of this work. By moving the gradient-based computation to the initial pre-training phase, CIFNet avoids the multiplicative cost that iterative CIL methods must pay: hundreds of epochs per task, repeated over long sequences of tasks, through millions of feature extractor parameters, not to mention the time involved in finding suitable hyperparameters. The result is a framework whose incremental training cost is dominated by a single forward pass and a matrix operation, regardless of task sequence length. We quantify this advantage in terms of training time, energy consumption, and estimated CO$_2$ emissions, establishing a systematic Green AI evaluation within the analytic CIL family.

We evaluate CIFNet on diverse visual datasets under standard class-incremental protocols with varying granularities. The results show that CIFNet achieves accuracy competitive with SGD-based CIL methods at a fraction of their computational cost, and performs comparably to exemplar-free analytic methods while introducing calibration for the class-expansion regime. Across all settings, CIFNet reduces energy consumption by up to $20\times$ relative to state-of-the-art iterative methods, without sacrificing performance stability.

The main contributions of this work are as follows:

\begin{itemize}
    \item We reformulate class-incremental learning as a sequence of closed-form convex problems on stationary representations, eliminating backbone fine-tuning and iterative gradient updates. 

    \item We formulate a closed-form Regularised Recursive Least-Squares (RRLS) classifier that accumulates sufficient statistics incrementally, guaranteeing exact global minimisation without convergence dynamics or learning rate tuning.

    \item We identify and resolve the structural initialisation bias in newly added output neurons using a compact embedding calibration buffer and density-aware oversampling, yielding globally balanced decision boundaries without raw image replay.

    \item We demonstrate through extensive benchmarks (CIFAR-100, ImageNet-100, CORe50) and a 50-task long-horizon sequence that CIFNet provides smooth, collapse-free learning trajectories while drastically reducing computational and energy footprints compared to iterative CIL paradigms.
\end{itemize}

The remainder of the paper is organised as follows. Section~\ref{related_work} situates CIFNet within the CIL literature, with particular attention to analytic and frozen-backbone methods. Section~\ref{subsec:problem} formalises the class-incremental learning setting. Section~\ref{sec:method} develops the three components of CIFNet in detail. Section~\ref{sec:evaluation} presents experimental results and sustainability analysis. Section~\ref{sec:conclusions} presents a summary of our findings, discusses limitations, and outlines directions for future work.

\section{Related Work}\label{related_work}

Continual Learning (CL) addresses the challenge of learning from a sequential data stream while retaining previous knowledge \cite{Parisi2019}. In Class Incremental Learning (CIL), new classes are introduced sequentially over time, requiring the model to incorporate them without access to the original training data while retaining knowledge of previously learned classes. The central obstacle is catastrophic forgetting, where gradient updates for new classes overwrite parameters critical for earlier ones \cite{Rebuffi2017}. Extensive surveys have documented the landscape of proposed solutions \cite{zhou2024cil_survey}, which generally fall into four families: regularisation and distillation, replay, dynamic architectures, and --more recently-- analytic closed-form methods. CIFNet is most closely related to the last family and to frozen-backbone replay approaches, and can be considered to fall explicitly within both.

\subsection{Regularisation and Distillation}

Regularisation-based methods constrain parameter updates to protect previously learned knowledge. Elastic Weight Consolidation (EWC) \cite{kirkpatrick2017overcoming} penalises changes to weights deemed important for past tasks via a Fisher-information penalty. Knowledge Distillation (KD) \cite{li2017learning} takes a complementary approach, regularising a student model to reproduce the output distribution of its predecessor. PODNet \cite{douillard2020podnet} extends this by distilling intermediate feature maps, reducing spatial forgetting across the network. Furthermore, TA \cite{szatkowski2024adapt} applies knowledge distillation directly in the embedding space to preserve representational structure. More recently, logit normalisation has been introduced to correct the recency bias that KD-based methods inherit when old and new class logits are on incomparable scales \cite{gao2025maintaining}, while EXACFS \cite{balasubramanian2024exacfs} selects class-wise feature activations for targeted preservation. A common limitation of this family is its sensitivity to hyperparameter choices---regularisation strengths, distillation temperatures, and learning rate schedules must all be carefully tuned for each incremental scenario.

\subsection{Replay Methods}

Replay methods maintain a small episodic buffer of past exemplars and interleave them with new task data during training \cite{Rebuffi2017, barry2023neural}. iCaRL \cite{Rebuffi2017} combines replay with nearest-class-mean classification and herding-based exemplar selection. Dark Experience Replay (DER) \cite{buzzega2020dark} additionally stores logit responses, providing a soft distillation signal from past model states. To reduce the storage cost of raw pixel buffers, exemplar-free variants like GILD \cite{jung2023generating} exploit frozen pre-trained representations, generating instance-level prompts from latent codes. 

\subsection{Dynamic Architectures}
Dynamic expansion avoids interference by expanding model capacity as new tasks arrive \cite{rusu2016progressive}. PackNet \cite{mallya2018packnet} prunes and freezes task-specific subnetworks, while FOSTER \cite{wang2022foster} and MEMO \cite{zhou2022model} compress task-specific feature boosters into a shared backbone. More recent approaches such as TagFex \cite{zheng2025task} expand representations dynamically while managing parameter growth through distillation and pruning. Parameter-efficient tuning methods, including prompt-based approaches for Vision Transformers such as L2P \cite{wang2022learning}, CODA-Prompt \cite{Smith_2023_CVPR}, HiDe-Prompt \cite{wang2023hierarchical}, and NoRGa \cite{le2024mixture}, have recently demonstrated compelling accuracy at modest parameter overhead, though they are inherently tied to transformer architectures and their associated computational infrastructure, which conflicts with settings with limited computational budgets. Finally, recent work such as SLCA \cite{zhang2023slca} and EASE \cite{zhou2024expandable} combine frozen or slowly adapting backbones with classifier alignment strategies, achieving strong accuracy at significantly reduced computational cost. Unlike the prior families, which often rely on complex, iterative optimisation schedules, these methods treat incremental learning as a sequence of adaptation steps, aligning the classification head to the evolving feature distribution.

\subsection{Analytic and Closed-Form Methods}
A smaller but growing body of work replaces iterative gradient descent with closed-form solutions, eliminating the hyperparameter sensitivity and multi-epoch overhead of SGD-based training. ACIL \cite{zhuang2022acil} establishes that class-incremental learning can be reformulated as a recursive least-squares problem, achieving mathematically exact weight updates with absolute resistance to forgetting without requiring an exemplar buffer. To improve class separability, ACIL performs classification in a high-dimensional latent space obtained through a fixed random projection of the backbone features before applying the analytic solver. DS-AL \cite{zhuang2024ds} extends this principle to a dual-stream architecture that separates semantic and structural feature components, while similarly relying on projected high-dimensional representations for closed-form learning. Building on similar principles, \cite{li2023CRNet} proposes a random-theory-based continual learning framework that leverages random feature mappings for efficient analytic updates.

In contrast, CIFNet performs analytic learning directly in the backbone embedding space without any feature expansion stage. Avoiding high-dimensional random projections reduces memory requirements and keeps covariance estimation computationally tractable. This represents a different operating point within the CIL design space, one that favours resource efficiency over representational expansion.

A key limitation shared by most analytic CIL methods is their assumption of a fixed or slowly evolving feature space: when applied with a frozen backbone, they rely on the quality of the pre-trained representation to cover all incremental classes. CIFNet operates in this same fixed-feature regime but addresses a distinct practical challenge these methods do not fully solve: the class imbalance introduced when new task data vastly outnumbers stored past representations. Through a calibration buffer of compact embeddings and density-aware oversampling, CIFNet corrects this distributional bias directly at the optimisation level, yielding balanced decision boundaries without storing raw images or performing any iterative updates.

\section{Problem Setting}
\label{subsec:problem}

CIL involves learning from a sequence of tasks $T_1, T_2, \ldots, T_K$, where each task introduces a disjoint set of novel classes. Tasks arrive sequentially, and once a task has been processed, its training data are no longer accessible. Formally, task $T_k$ provides a dataset $\mathcal{D}_k = {(x_i, y_i)}_{i=1}^{N_k}$. While CIL is modality-agnostic, this work focuses on visual recognition, where each input image is represented as $x_i \in \mathbb{R}^{h \times w \times c}$, with $h$, $w$, and $c$ denoting its height, width, and number of channels, respectively. Labels $y_i$ are drawn from a task-specific class set $\mathcal{C}_k$. Tasks are disjoint, such that $\mathcal{C}_i \cap \mathcal{C}_j = \emptyset$ for $i \neq j$. At inference, the model must classify samples from the cumulative label space $\mathcal{Y}_k = \bigcup_{i=1}^k \mathcal{C}_i$ without being provided the task identity.

The core difficulty is \textit{catastrophic forgetting}: when a neural network is updated via SGD on $\mathcal{D}_k$, gradient steps that minimise loss on new classes systematically overwrite parameters encoding knowledge of earlier ones \cite{McCloskey1989}. Replay-based methods mitigate this by maintaining a buffer $B_{k-1}$ of exemplars from past tasks and jointly optimising over $\mathcal{D}_k \cup B_{k-1}$. However, even with replay, iterative methods remain sensitive to learning rate schedules, epoch counts, and the relative sizes of current data and buffer, and they require repeated backpropagation through the full network at every incremental step, compounding both computational and environmental cost.

CIFNet bypasses these difficulties by restricting gradient-based learning entirely to the pre-training phase. During incremental learning, only the classifier is updated, and it is updated analytically.

\section{CIFNet: A Frugal Analytic Framework for Class-Incremental Learning}
\label{sec:method}

\begin{figure*}[t]
    \centering
    \includegraphics[width=\linewidth]{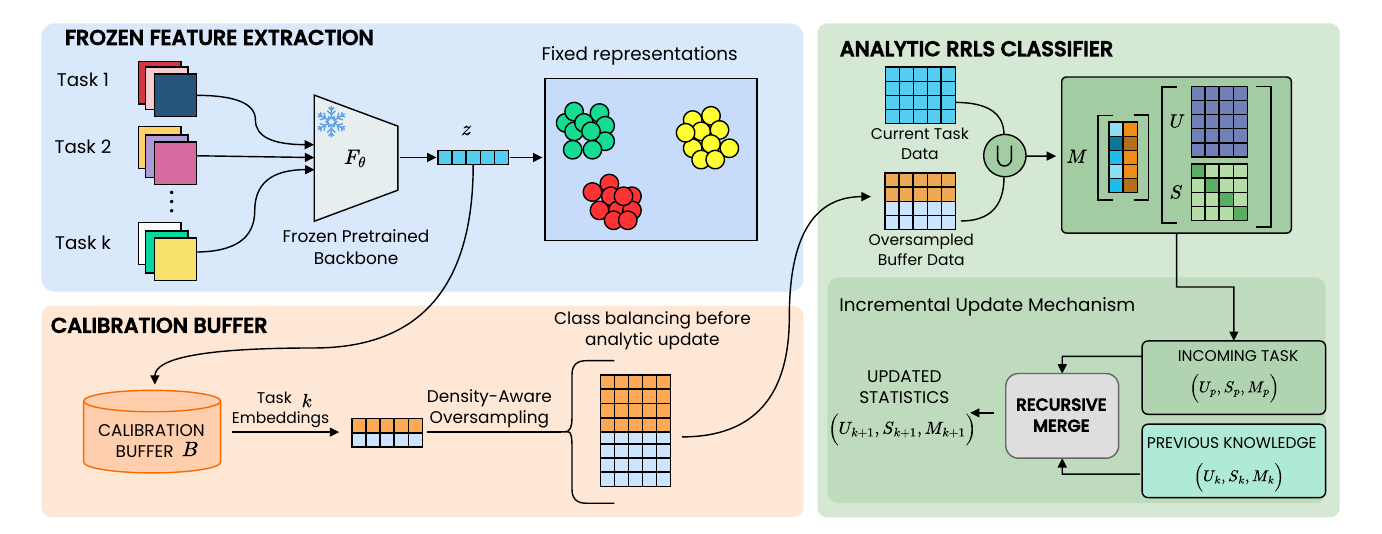}
    \caption{\textbf{CIFNet schematic diagram.} Incremental learning occurs entirely in the classifier space while the embedding space remains fixed. \textbf{Blue (Top):} A frozen pre-trained backbone projects incremental tasks into a stationary latent space ($z \in \mathbb{R}^d$). \textbf{Orange (Bottom):} A lightweight calibration buffer stores compact embeddings for density-aware oversampling. Buffer embeddings are applied only to balance newly introduced classes ($\mathcal{C}_k$), while previous class weights are locked. \textbf{Green (Right):} Fused embeddings undergo an analytic three-stage RRLS closed-form solution. The incremental update mechanism expands the classifier by recursively merging previous statistics $(U_k, S_k, M_k)$ and incoming task statistics $(U_p, S_p, M_p)$.}
    \label{fig:method}
\end{figure*}

As motivated in the previous section, CIFNet eliminates iterative optimisation during incremental learning by freezing the feature extractor after pre-training and performing all subsequent updates analytically in the classifier space. This design transforms class-incremental learning into a sequence of deterministic linear updates, avoiding backpropagation, learning-rate schedules, and optimiser state while substantially reducing computational cost. Beyond competitive accuracy, this design substantially reduces the computational cost of continual learning by eliminating backward passes and optimiser state updates, reducing both training time and energy consumption.

Figure~\ref{fig:method} provides an overview of the proposed CIFNet framework. The method comprises three interlocking components: (i) a frozen pre-trained backbone that projects all incoming samples into a stationary embedding space (Section~\ref{subsec:backbone}), (ii) an analytic RRLS classifier that incrementally updates the decision space through deterministic closed-form solutions (Section~\ref{subsec:rrls}), and (iii) a lightweight calibration buffer that mitigates class imbalance during incremental expansion (Section~\ref{subsec:buffer}).

\subsection{Stationary Embedding Space via Frozen Backbone}
\label{subsec:backbone}

A pre-trained network $F_\theta$ maps each input image $x \in \mathbb{R}^{ h \times w \times c} $ to a compact embedding:

\begin{equation}
    z = F_\theta(x) \in \mathbb{R}^d, \qquad 
    d \ll h \times w \times c.
\end{equation}

By freezing parameters $\theta$ throughout incremental training, the geometry of the embedding space is held constant across all tasks. This has two important consequences. First, representational drift is structurally avoided, meaning that the relative positions of previously learned class embeddings remain fixed throughout the incremental sequence. This preserves earlier decision boundaries when new classes arrive. Second, feature extraction is reduced to a single forward pass per sample (just to obtain the embedding), removing the need for gradient computation. The resulting embeddings can then be cached and reused by the classifier.

This design assumes that the backbone produces sufficiently transferable features across all incremental classes, which depends on compatibility between pre-training and deployment distributions. While this assumption is increasingly reasonable for large-scale pre-trained models \cite{zhou2025revisiting}, it remains a limitation that may affect performance under strong domain shift. We discuss this limitation explicitly in Section~\ref{sec:conclusions}.

\subsection{Closed-Form Incremental Classifier}
\label{subsec:rrls}

With the embedding space fixed, in this framework, incremental learning reduces to a sequence of linear classification problems. Instead of iterative gradient descent, which introduces sensitivity to optimisation schedules and increases energy consumption due to repeated backward passes, CIFNet computes classifier weights via a deterministic, single-step Regularised Recursive Least-Squares (RRLS) formulation.

While the RRLS layer is mathematically grounded in previous analytic learning theory~\cite{fontenla2021rolann}, we employ it here as a deterministic, closed-form ridge-regression solver operating on fixed representations. Applying RRLS to the class-incremental setting, however, requires accommodating the progressive expansion of the classifier as new classes are introduced. This setting gives rise to structural initialisation bias in newly added output neurons, motivating the calibration mechanism described in Section~\ref{subsec:buffer}. Overall, the RRLS layer decouples classifier optimisation from feature learning, removing variability associated with initialisation and training dynamics.

\subsubsection{Problem Formulation}

Let $Z \in \mathbb{R}^{n \times d}$ denote the embedding matrix for $n$ samples from the current task, and let $Y \in \{0,1\}^{n \times |\mathcal{C}|}$ be the one-hot label matrix for the set of seen classes $\mathcal{C}$. To obtain a stable regression target in a linear framework, we transform labels into a \emph{pre-activation space} via a clipped inverse-sigmoid mapping:
\begin{equation}
    \hat{Y} = f^{-1}(Y) \in \mathbb{R}^{n \times |\mathcal{C}|}
    \label{eq:delinear}
\end{equation}

\noindent where $f^{-1}$ denotes the logit function applied to probabilities clipped to the range $[\epsilon, 1-\epsilon]$, with $\epsilon \ll 1$ being a small positive constant to ensure numerical stability. 

While this framework naturally scales to compute simultaneously the weights for every class via matrix operations, for mathematical clarity we present the following closed-form derivation for a single output neuron (i.e., one class). Let $\hat{y} \in \mathbb{R}^n$ denote the column vector from $\hat{Y}$ corresponding to the specific class under consideration. In practice, all stages described below are computed independently and in parallel across all $\mathcal{C}$ classes.

\subsubsection{Closed-Form Solution via SVD}

The weight computation proceeds in three stages that together yield the ridge regression normal equations in the principal subspace of a derivative-weighted feature matrix.

\paragraph{Stage 1: Feature construction.}

We augment embeddings with a bias term:
\begin{equation}
    X = \begin{bmatrix} \mathbf{1} & Z \end{bmatrix}^{\!\top} \in \mathbb{R}^{(d+1) \times n}
\end{equation}

Fitting the weights to the pre-activation targets $\hat{y}$ requires accounting for the nonlinear relationship between the classifier output and the pre-activation. A first-order linearisation of the chosen activation function around the target values introduces a per-sample weighting through its derivative: samples whose targets correspond to operating points where the sigmoid is steep contribute more to the gradient of the least-squares objective than samples in the saturated regime. Concretely, the diagonal weighting matrix is
\begin{equation}
    F = \mathrm{diag}\!\left(f'(\hat{y}_1), \ldots, f'(\hat{y}_n)\right) \in \mathbb{R}^{n \times n}
\end{equation}

\noindent Here, $f$ denotes the invertible activation function used to map the unbounded pre-activation space to the probability domain. In this work we employ the sigmoid function, whose derivative is $f'(x) = f(x)(1-f(x))$, and $\hat{y}_i$ is the target scalar for sample $i$. Each diagonal entry modulates the corresponding sample's contribution to the subsequent computation. The derivative-weighted feature matrix is:
\begin{equation}
    H = X F \in \mathbb{R}^{(d+1) \times n}
\end{equation}

\paragraph{Stage 2: SVD decomposition and moment accumulation.}

A reduced Singular Value Decomposition (SVD) is then applied to $H$:
\begin{equation}
    H = U S V^{\top}
    \label{eq:svd_rrls}
\end{equation}

\noindent where $U \in \mathbb{R}^{(d+1) \times r}$ contains the left singular vectors and $S \in \mathbb{R}^{r \times r}$ is a diagonal matrix containing the corresponding singular values, with $r = \min(d+1, n)$. The columns of $U$ form an orthonormal basis for the principal directions of the weighted feature space, while the singular values quantify the variance captured along each direction. This decomposition provides a numerically stable basis for the regularised inversion performed in the next stage.

In parallel with the SVD, a moment vector $m \in \mathbb{R}^{d+1}$ is computed to capture the correlation between the weighted features and the pre-activation targets for the corresponding class:
\begin{equation}
    m = X F^\top F \hat{y}
    \label{eq:moment_rrls}
\end{equation}

Geometrically, $m$ accumulates the class-specific evidence provided by each feature dimension, with each sample weighted by the squared derivative $F^\top F$. This weighting follows directly from the normal equations of the weighted least-squares problem and is therefore consistent with the linearisation introduced in the previous stage. Beyond its role in the current optimisation, $m$ possesses an important structural property that enables the recursive formulated introduced in Section~\ref{sub:recursive}: it is additive over data partitions. If a second batch yields moment vector $m'$, the combined moment is simply $m + m'$, without revisiting previously observed samples.

\paragraph{Stage 3: Regularised Weight Solution.}

Given the SVD of $H$ and the moment vector $m$, the classifier weight for a single neuron is obtained as
\begin{equation}
    w = U (S^2 + \lambda I)^{-1} U^{\top} m
    \label{eq:weights_rrls}
\end{equation}

\noindent where $\lambda > 0$ is the ridge regularisation parameter. This solution performs ridge regression in the principal subspace of $H$: each singular direction $u_i$ contributes proportionally to $s_i^2/(s_i^2+\lambda)$. Consequently, well-supported directions (large singular values) are preserved almost unchanged, whereas poorly supported directions are progressively attenuated, improving numerical stability and reducing the influence of noise.

Since the optimisation is a strictly convex ridge regression problem on fixed features, Equation~\eqref{eq:weights_rrls} is the unique global optimum for any $\lambda>0$. Consequently, the resulting classifier is deterministic: identical feature representations always produce identical classifier parameters, eliminating the variability associated with parameter initialisation, stochastic optimisation, and learning-rate schedules. A formal proof of the uniqueness of the solution is provided in Appendix~\ref{app:rrls-proof}.

\subsubsection{Recursive Incremental Update}
\label{sub:recursive}

Having established the closed-form solution for a single neuron, we now define the global state of the classifier layer. For a network with $|\mathcal{C}|$ classes, the global parameters are maintained as collections of the class-specific statistics: the moment matrix $M \in \mathbb{R}^{(d+1) \times |\mathcal{C}|}$ (where each column is a class vector $m_c$), the weight matrix $W \in \mathbb{R}^{(d+1) \times |\mathcal{C}|}$, and the subspace bases $U = \{U_1, \dots, U_C\}$ and singular values $S = \{S_1, \dots, S_C\}$.

The property that makes the RRLS layer central to CIFNet is the additive nature of these sufficient statistics. For a new data partition $p$, its multi-class statistics $(U_p, S_p, M_p)$ can be merged with the previously accumulated knowledge $(U_k, S_k, M_k)$. The global moment matrix accumulates additively:
\begin{equation}
    M_{k+1} = M_k + M_p
    \label{eq:moment_update}
\end{equation}

 To maintain a compact representation, the principal subspaces can be updated for each class $c \in \mathcal{C}$ via:
\begin{equation}
    \bigl[U_{c, k+1},\, S_{c, k+1},\, \sim\bigr] = 
    \mathrm{SVD}\bigl([U_{c,k} S_{c,k} \;\|\; U_{c,p} S_{c,p}]\bigr)
    \label{eq:svd_update}
\end{equation}

The concatenated matrix spans the union of the two principal subspaces weighted by their respective singular values, and the reduced SVD projects this union back onto an orthonormal basis of fixed rank---a numerically well-studied subspace merging operation \cite{brand2006fast}. The updated global weight matrix $W_{k+1}$ follows immediately from computing Equation~\eqref{eq:weights_rrls} in parallel across all merged class statistics. This structure means that each incremental step requires only the statistics of the incoming partition and the accumulated summary $(U_k, S_k, M_k)$, never the original data.
Forgetting is avoided by design at the classifier level, Equation~\eqref{eq:moment_update} is equivalent to computing $M$ on the union of all observed partitions, and the SVD merge in Equation~\eqref{eq:svd_update} preserves the corresponding principal subspace. In practice, this yields an efficient approximation to training on the full union of past data under a fixed-rank constraint, without revisiting raw samples.

Unlike \cite{zhuang2022acil} and \cite{zhuang2024ds}, which introduce an intermediate feature expansion stage before the analytic solver, CIFNet operates directly on native backbone embeddings. As a result, the memory footprint of the analytic component scales with the original feature dimensionality rather than an expanded projection space. This significantly decreases the memory required to store sufficient statistics for the RRLS update. For each class $c$, the RRLS layer maintains its matrix $M$, its specific basis matrix $U_c$, singular values $S_c$, and the weight vector $w_c$, where the rank $r = \min(d+1, n)$ is bounded by the feature dimension.

In the CIL setting, each new task introduces previously unseen classes. CIFNet handles this by dynamically expanding the topology of the classifier: for task $k$, $|\mathcal{C}_k|$ new output neurons are added to $W$. Each new neuron is introduced without accumulated sufficient statistics for its corresponding class. These statistics are initialised from the first observations of that class and updated recursively thereafter, meaning it has no prior exposure to features from previous classes. Because the RRLS solution for each class weight vector is derived from its own statistics but evaluated in a shared, fixed embedding space, the weights for previously learned classes are preserved exactly when the embedding space is fixed. However, this creates a `cold start' problem: the new neurons are trained exclusively on current-task data, making them prone to structural initialisation bias. Without past-class evidence, the new neurons cannot learn to suppress activation on previously learned embedding regions. We solved this problem by introducing a calibration buffer: it provides the only source of past-class evidence available to these new neurons, allowing them to define decision boundaries that are globally consistent across both new and old classes.

\subsection{Calibration Buffer and Density-Aware Oversampling}
\label{subsec:buffer}

\begin{figure}[t]
    \centering
    \includegraphics[width=0.95\linewidth]{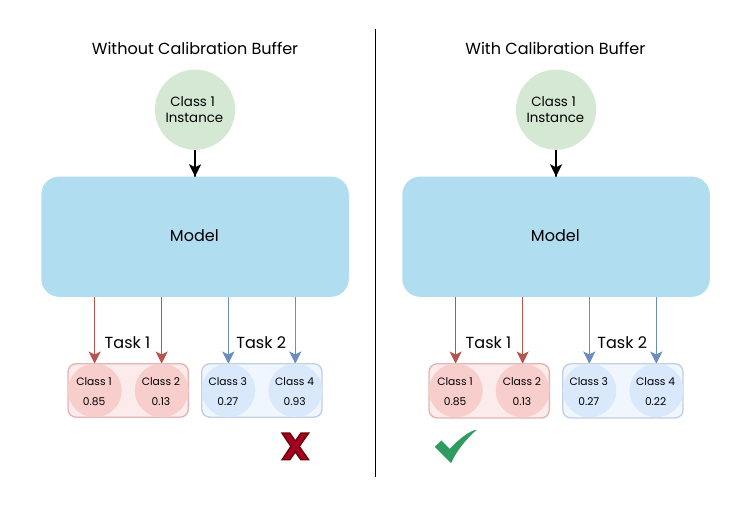}
    \caption{Effect of neuron calibration. Without calibration (left), neurons introduced in Task~2 strongly activate on Task~1 embeddings, producing biased predictions. With calibration (right), the analytic solution suppresses activation on negative class embeddings, yielding balanced decision boundaries across all seen classes.}
    \label{fig:buffer_overlapping}
\end{figure}

The recursive update and neuron-level isolation described above guarantee that the weights associated with neurons for classes $\mathcal{C}_{1:k-1}$ are not modified. However, the newly added neurons for $\mathcal{C}_k$ are trained exclusively on current task data and therefore have no exposure to embeddings from past classes. Without corrective information, these neurons may learn to activate strongly on past-class regions of the embedding space; this occurs not because the old neurons have forgotten their original knowledge, but because the new ones lack exposure to those latent regions. This is a structural initialisation bias, distinct from catastrophic forgetting, that requires a swifter mechanism to address it effectively.

\subsubsection{Calibration Buffer}

CIFNet maintains a Calibration Buffer $B$ of stored embeddings from past tasks, whose purpose is to provide the newly added neurons with evidence from previously learned classes. At task $k$, the buffer data $B_{k-1}$ are presented only to the $|\mathcal{C}_k|$ new neurons during their RRLS update; neurons for classes $\mathcal{C}_{1:k-1}$ are not exposed to the buffer again. This use of the buffer is both computationally efficient and algebraically consistent; it does not alter any previously computed solution and adds no new forward passes to the backbone.

Unlike standard replay buffers, which retain raw images for iterative re-training of the full network, $B$ operates entirely in the latent space $\mathbb{R}^d$ and serves the specific algebraic role of suppressing excessive activation of new neurons on past-class regions of the feature space (Figure~\ref{fig:buffer_overlapping}). Operating in embedding space rather than pixel space yields a substantial practical benefit. For example, with a ResNet-18 backbone with $d = 512$, storing 2,000 embeddings as float32 vectors requires approximately $4.1$\,MB, compared to $1{,}206$\,MB for the equivalent number of raw ImageNet-scale images, a reduction of nearly $300\times$ that enables dense coverage of past class distributions under tight memory budgets.

The buffer is populated via random sampling rather than herding or distance-based selection. This sampling strategy is chosen since it preserves the empirical class distribution without introducing additional selection criteria. As shown in Section~\ref{subsec:ablation}, it was observed experimentally that this strategy produced the most stable performance among the alternatives evaluated.

\subsubsection{Density-Aware Oversampling}

Even with a calibration buffer, the optimisation in Equation~\eqref{eq:objective_rrls} remains unbalanced: the current task contributes $N_k$ training samples, while each past class is represented by at most $m$ stored embeddings, where typically $m \ll N_k$. Left uncorrected, this imbalance causes the moment matrix $M$ and the SVD in Equation~\ref{eq:svd_update} to be dominated by current-task statistics, weakening the calibration effect that the buffer is intended to enforce.

To correct this, we apply density-aware oversampling to the stored past-class embeddings during the classifier update. Rather than physically replicating data in the buffer, we increase the relative contribution of each past class $c$ to the RRLS objective by a factor:
\begin{equation}
    r_c = \left\lfloor \frac{n_{\max}}{n_c'} \right\rfloor,
    \label{eq:oversampling}
\end{equation}

\noindent where $n_{\max} = \max_{j \in \mathcal{C}_k} N_j$ is the most widely represented class in the current task and $n_c'$ is the number of stored samples for class $c$. This strategy effectively balances the weights of past and current classes in the RRLS estimation without increasing the memory footprint of the buffer.

The complete procedure is summarised in Algorithm~\ref{alg:training_phase}.

\begin{algorithm}[t]
    \caption{CIFNet Incremental Training}
    \label{alg:training_phase}
    \textbf{Input:} Training data $\mathcal{D}_k$; frozen extractor 
    $F_\theta$; classifier $G_{\phi_{k-1}}$; calibration buffer 
    $B_{k-1}$; per-class buffer capacity $m$ \\
    \textbf{Output:} Updated classifier $G_{\phi_k}$; updated 
    buffer $B_k$
    \begin{algorithmic}[1]
        \Function{IncrementalTrain}{$\mathcal{D}_k, F_\theta, 
            G_{\phi_{k-1}}, B_{k-1}, m$}
            \State Expand $G_{\phi_{k-1}}$: add $|\mathcal{C}_k|$ output neurons
            \State $Z_k \leftarrow \{(F_\theta(x_i), y_i) \mid (x_i,y_i) \in \mathcal{D}_k\}$
            \State $\tilde{B} \leftarrow$ 
                \Call{OversampleBuffer}{$Z_k, B_{k-1}$}
                \Comment{Eq.~\eqref{eq:oversampling}, new neurons only}
            \State $X_k, Y_k \leftarrow Z_k \cup \tilde{B}$
            \State $G_{\phi_k} \leftarrow$ \Call{TrainRRLS}{$X_k, Y_k, G_{\phi_{k-1}}$}
                \Comment{Eqs.~\eqref{eq:delinear}--\eqref{eq:weights_rrls} and \eqref{eq:moment_update}--\eqref{eq:svd_update}}
            \State $B_k \leftarrow$ \Call{UpdateBuffer}{$Z_k, B_{k-1}, m$}
            \State \Return $G_{\phi_k}, B_k$
        \EndFunction
    \end{algorithmic}
\end{algorithm}

\section{Experimental Evaluation}
\label{sec:evaluation}

In this section, we provide a rigorous analysis of the experimental results, focusing on accuracy, learning stability, and computational efficiency, rather than isolated performance figures. The evaluation is organised around three findings. First, operating on fixed pre-trained representations is sufficient to match or exceed the accuracy of iterative backbone-adaptive methods at a substantially lower computational cost. Second, within the frozen-backbone regime, there exists a critical trade-off between the stability of the classifier's update mechanism and its capacity to adapt to class-specific feature distributions; standard benchmarks often obscure this by favouring either high-variance iterative updates or overly conservative analytic constraints. Finally, the energy and memory footprint of analytic methods vary significantly depending on the dimensionality of the feature space in which classification is performed, with CIFNet occupying the most resource-efficient operating point across all evaluated configurations.

\paragraph{Datasets and Metrics}

We evaluate the proposed method on three standard CIL benchmarks: CIFAR-100 \cite{krizhevsky2009learning}, ImageNet-100 \cite{le2015tiny}, and CORe50 \cite{lomonaco2017core50}. CIFAR-100 contains 50,000 training and 10,000 test images across 100 classes. ImageNet-100 is a 100-class subset of ImageNet-1k with 130,000 training images and 5,000 validation images. CORe50 is designed for continual object recognition in realistic robotic settings, comprising 50 objects observed under 11 sequential video-recorded sessions with significant environmental variation. Due to the high temporal redundancy inherent in the original video recordings, we subsample every 10th frame, yielding 16,486 images and ensuring greater visual diversity across sessions.

Performance is measured using the final cumulative accuracy $A_K$, defined as the accuracy obtained after learning the last task and evaluated over all classes seen throughout the incremental sequence. We also report the average accuracy $\bar{A} = \frac{1}{K}\sum_{k=1}^{K}A_k$, where $A_k$ denotes the cumulative accuracy after learning task $k$, evaluated over all classes in $\mathcal{C}_{1:k}$. This metric captures stability and performance across the entire incremental learning trajectory. In addition, we assess computational and environmental efficiency using sustainability metrics. Utilising CodeCarbon \cite{codecarbon}, we track total execution runtime (measured in minutes) and overall energy consumption (measured in kilowatt-hours, kWh).

\paragraph{Selected Methods and Evaluation Protocol}

We compare CIFNet against representative CIL methods spanning the main learning paradigms, including backbone-adaptive methods trained from random initialisation (iCaRL \cite{Rebuffi2017}, PODNet \cite{douillard2020podnet}, WA \cite{wu2019large}, DER \cite{buzzega2020dark}, RMM-FOSTER \cite{wang2022foster}, MEMO \cite{zhou2022model}, and CREATE \cite{chen2025reducing}), backbone-adaptive methods with pre-trained initialisation (TagFex \cite{zheng2025task}), and frozen-backbone analytic methods operating on fixed pre-trained representations (ACIL \cite{zhuang2022acil}, DS-AL \cite{zhuang2024ds}, and CIFNet), alongside a standard Finetune baseline. To prioritise frugal, resource-constrained learning, we restrict our evaluation to CNN-based architectures, explicitly excluding transformer-based approaches (e.g., L2P \cite{wang2022learning} and CODA-Prompt \cite{Smith_2023_CVPR}) due to their substantially higher computational and memory requirements.

This grouping is maintained throughout the analysis to ensure that comparisons reflect different continual learning strategies rather than implementation artefacts. In particular, differences in accuracy, trajectory stability, and computational efficiency are interpreted within the context of each operating regime rather than assuming that all methods optimise for the same objective.

All methods relying on pre-trained backbones use self-supervised rather than supervised pre-training, avoiding potential label leakage from ImageNet-1k into the ImageNet-100 benchmark while ensuring that performance differences reflect the quality of the incremental learning strategy rather than prior label supervision. All experiments are repeated using three different random seeds, and reported results correspond to the average across runs.

\paragraph{Implementation Details}

All experiments use a ResNet-18 \cite{he2016deep} backbone as the feature extractor, a standard architecture widely adopted in the CIL literature that provides a balanced trade-off between representational capacity and computational cost. All buffer-based methods are constrained to 2,000 stored samples. Traditional replay and distillation approaches store raw images, requiring approximately $1,206.40$\,MB on ImageNet-100, whereas CIFNet stores compact latent embeddings, reducing the buffer footprint to $4.10$\,MB under the same exemplar constraint.

CIFNet is implemented in PyTorch \footnote{The link to the code on GitHub will be left here if the article is accepted.}.  Regarding training hyperparameters, CIFNet requires no learning rate, epoch count, or optimiser schedule. All other baselines follow configurations from their original implementations or \cite{zhou2024cil_survey}. Experiments were conducted on an NVIDIA RTX 3090 Ti GPU and an Intel Core i7-12700K CPU.

\subsection{Overall Performance}
\label{subsec:accuracy}

\begin{table*}[!ht]
\caption{Accuracy comparison on CIFAR-100, ImageNet-100, and CORe50 across different class-incremental settings. Best and second-best results are shown in bold and underlined, respectively.}
\label{tab:accuracy-results}
\centering
\resizebox{\linewidth}{!}{%
\begin{tabular}{lcccccccccccccc}
\toprule
& \multicolumn{4}{c}{CIFAR-100} & \multicolumn{4}{c}{ImageNet-100} & \multicolumn{4}{c}{CORe50} \\
\cmidrule(lr){2-5} \cmidrule(lr){6-9} \cmidrule(lr){10-13}
Model 
& \multicolumn{2}{c}{5 cls/task}& \multicolumn{2}{c}{10 cls/task}& \multicolumn{2}{c}{5 cls/task}& \multicolumn{2}{c}{10 cls/task}& \multicolumn{2}{c}{5 cls/task}& \multicolumn{2}{c}{10 cls/task} \\
\cmidrule(lr){2-3}\cmidrule(lr){4-5}\cmidrule(lr){6-7}\cmidrule(lr){8-9}\cmidrule(lr){10-11}\cmidrule(lr){12-13}
& $\bar{A}$ & $A_K$& $\bar{A}$ & $A_K$& $\bar{A}$ & $A_K$& $\bar{A}$ & $A_K$& $\bar{A}$ & $A_K$& $\bar{A}$ & $A_K$ \\
    \midrule
    Finetune & 17.58 & 5.25 & 28.18 & 9.02 & 17.60 & 4.64 & 28.40 & 9.34 & 21.70 & 8.50 & 33.23 & 15.80 \\
    \midrule \rowcolor{gray!10}\multicolumn{13}{l}{\textit{Backbone-adaptive (random initialization)}}\\
    iCaRL & 54.58 & 34.87 & 61.51 & 40.76 & 54.97 & 33.06 & 61.59 & 40.18 & 56.61 & 44.40 & 58.09 & 45.83 \\
    PODnet & 49.22 & 28.41 & 57.52 & 37.86 & 55.83 & 37.58 & 65.80 & 46.78 & 46.30 & 31.13 & 51.13 & 35.20 \\
    RMM-FOSTER & 67.03 & 51.10 & 70.86 & 57.12 & 71.73 & 59.46 & 76.49 & 66.10 & 57.55 & 55.67 & 62.36 & 58.87 \\
    WA & 59.30 & 42.75 & 66.97 & 51.90 & 63.21 & 46.72 & 70.92 & 55.44 & 54.98 & 47.83 & 59.12 & 52.20 \\
    CREATE & 66.25 & 47.60 & 70.01 & 56.26 & 69.50 & 53.88 & 77.11 & 65.36 & 43.37 & 31.47 & 48.58 & 39.77 \\
    MEMO & 68.49 & 54.34 & 71.99 & 58.46 & 68.42 & 55.52 & 72.66 & 60.94 & 61.51 & 56.00 & 61.66 & 56.23 \\
    DER & 71.34 & 57.34 & 70.80 & 59.03 & 73.88 & 63.66 & 76.04 & 64.40 & 59.27 & 52.27 & 59.60 & 54.30 \\ 
    \midrule \rowcolor{gray!10}\multicolumn{13}{l}{\textit{Backbone-adaptive (pre-trained)}}\\
    TagFex & \textbf{77.28} & \textbf{67.83} & \textbf{77.64} & \textbf{69.30} & 79.08 & 71.34 & 81.45 & 73.32 & 47.70 & 51.2 & 48.75 & 52.53 \\
    \midrule \rowcolor{gray!10}\multicolumn{13}{l}{\textit{Frozen backbone (analytic)}}\\
    ACIL & 71.59 & 66.76 & 72.14 & 67.11 & \textbf{91.21} & \underline{88.74} & \underline{90.83} & \underline{88.48} & 76.60 & \textbf{83.61} & \underline{78.38} & \textbf{83.65}  \\
    DS-AL & 71.51 & \underline{67.05} & 72.42 & \underline{67.43} & \underline{91.19} & \textbf{88.86} & \textbf{91.03} & \textbf{88.82} & \underline{77.02} & \underline{83.58} & 77.76 & \underline{83.62} \\
    \midrule
    CIFNet (Ours) & \underline{73.56} & 59.26 & \underline{72.93} & 59.57 & 87.87 & 81.10 & 87.73 & 81.04 & \textbf{77.18} & 76.91 & \textbf{78.51} & 77.53 \\
\bottomrule
\end{tabular}
}
\end{table*}

Table~\ref{tab:accuracy-results} reports the cumulative average accuracy ($\bar{A}$) and final accuracy ($A_K$) across all three benchmarks under 5- and 10-class incremental settings. Several observations can be drawn. 

In terms of $\bar{A}$, frozen-backbone analytic methods (ACIL, DS-AL, and CIFNet), remain competitive with backbone-adaptive approaches despite requiring no incremental backbone optimisation. On CIFAR-100, TagFex, which combines pre-trained initialisation with continued backbone adaptation, achieves the highest $\bar{A}$. On ImageNet-100, the best results are obtained by ACIL and DS-AL, while on CORe50, CIFNet achieves the highest $\bar{A}$ under both incremental settings.

Considering $A_K$, ACIL and DS-AL consistently outperform CIFNet on CIFAR-100 and ImageNet-100, whereas TagFex achieves the highest $A_K$ on CIFAR-100 among all evaluated methods. Among backbone-adaptive approaches, older replay-based methods such as iCaRL and PODNet are consistently surpassed by more recent approaches, including DER, MEMO, and RMM-FOSTER, across all datasets and task granularities.

CORe50 presents a markedly different ranking from the other benchmarks. While several methods that perform strongly on CIFAR-100 and ImageNet-100 experience noticeable reductions in $\bar{A}$ on CORe50, CIFNet achieves the highest average accuracy in both incremental settings. The implications of this behaviour are analysed in the following subsections.

\subsection{Learning Trajectory}
\label{subsec:trajectory}

\begin{figure*}
    \centering
    \includegraphics[width=\textwidth]{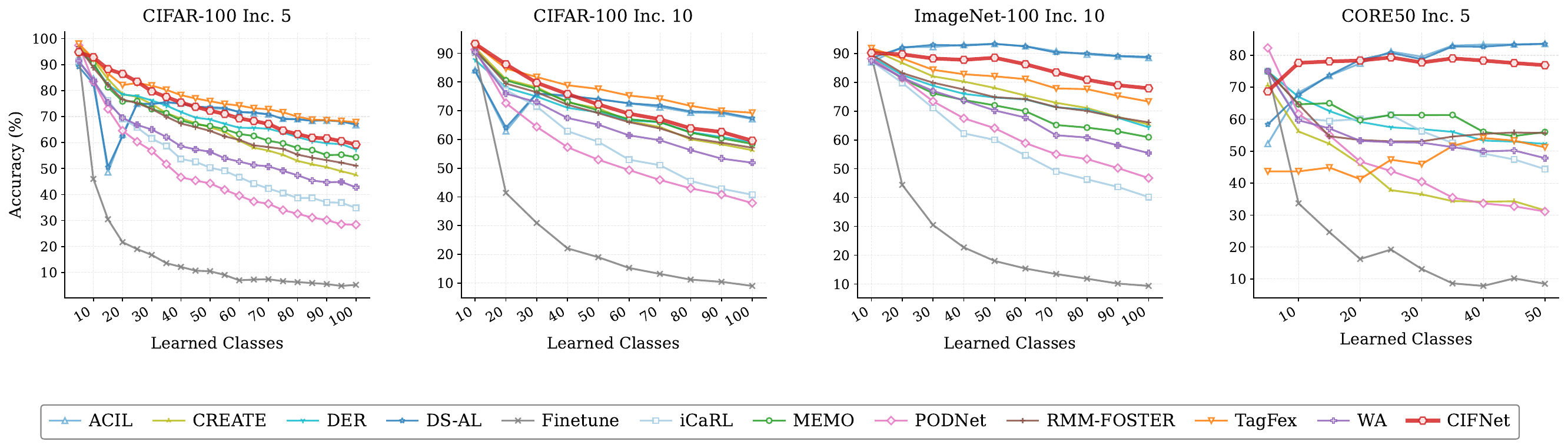}
    \caption{Incremental learning performance ($A_k$) on selected class-increment settings; representative configurations are shown for clarity.}
    \label{fig:curves}
\end{figure*}

Accuracy at the final task $A_K$ captures only the endpoint of a sequential learning process. For systems that are queried continuously throughout deployment (not only after all tasks have been observed), the shape of the accuracy trajectory across intermediate tasks is equally important. Figure~\ref{fig:curves} provides this view for all datasets.

Backbone-adaptive iterative methods generally exhibit a progressive decline in accuracy as more tasks are incorporated, reflecting the accumulation of representation drift and classifier interference over time. CIFNet, by contrast, follows a smooth and monotonic degradation. Because the embedding space remains fixed, no individual task update can substantially alter previously established decision boundaries, and the observed performance decay is governed primarily by the growing complexity of the classification problem rather than by representational instability.

The trajectory differences observed in Figure~\ref{fig:curves} explain the relationship between $\bar{A}$ and $A_K$ in Table~\ref{tab:accuracy-results}. Methods exhibiting abrupt fluctuations during the incremental sequence tend to present larger gaps between both metrics, whereas smoother trajectories produce more consistent values.

Among the analytic methods, ACIL and DS-AL exhibit noticeable drops in intermediate-task accuracy, particularly on CIFAR-100, before partially recovering in later tasks. Such transient performance drops may be undesirable in continual deployment scenarios, where the model is expected to provide reliable predictions throughout the learning process rather than only after the final task. By contrast, CIFNet maintains a smooth, monotonically decreasing trajectory throughout the incremental sequence, avoiding sudden regressions and providing more predictable behaviour during continual deployment.

\subsection{Computational Efficiency}
\label{subsec:efficiency}

\begin{figure}[t]
  \includegraphics[width=0.95\linewidth]{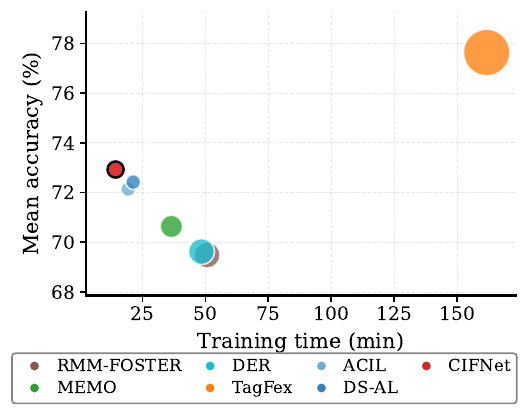}
  \caption{Mean accuracy ($\bar{A}$) versus training time on CIFAR-100 with 10 classes per task. Bubble size is proportional to energy consumed (kWh). Methods in the upper-left region achieve higher accuracy with shorter training times. CIFNet lies on the Pareto-optimal frontier among methods with comparable energy consumption.}
  \label{fig:bubble_green}
\end{figure}

\begin{table}[!htb]
\caption{Accuracy–efficiency trade-off $A_K$/kWh on CIFAR-100 (10 classes/task). Models are selected based on Pareto efficiency, balancing competitive accuracy with computational overhead. Downward ($\downarrow$) and upward ($\uparrow$) arrows denote metrics where lower and higher values are better, respectively.}
\label{tab:cifar100-efficiency}
\centering
\small
\resizebox{\columnwidth}{!}{
\begin{tabular}{lccccc}
    \toprule
    Model & Params (M) $\downarrow$ & Energy (kWh) $\downarrow$ & $A_K$ (\%) $\uparrow$ & $A_K$/kWh $\uparrow$ \\
    \midrule
    RMM-FOSTER & \textbf{0.46} & 0.298 & 57.12 & 191.7 \\
    MEMO & 7.14 & 0.202 & 58.46 & 289.4 \\
    DER & 9.27 & 0.302 & 59.03 & 195.5 \\
    TagFex & 7.42 & 1.151 & \textbf{69.30} & 60.2 \\
    ACIL & \textbf{0.46} & 0.081 & 67.11 & 828.51 \\
    DS-AL & \textbf{0.46} & 0.089 & 67.43 & 757.64 \\
    \midrule
    CIFNet (Ours) & \textbf{0.46} & \textbf{0.060} & 59.57 & \textbf{992.8} \\
    \bottomrule
\end{tabular}
}
\end{table}

\begin{table}[!htb]
\caption{Accuracy–efficiency trade-off $A_K$/kWh on ImageNet-100 (10 classes/task). Models are selected based on Pareto efficiency, balancing competitive accuracy with computational overhead. Downward ($\downarrow$) and upward ($\uparrow$) arrows denote metrics where lower and higher values are better, respectively.}
\label{tab:imagenet100-efficiency}
\centering
\small
\resizebox{\columnwidth}{!}{
\begin{tabular}{lcccc}
\toprule
Model & Params (M) $\downarrow$ & Energy (kWh) $\downarrow$ & $A_K$ (\%) $\uparrow$ & $A_K$/kWh $\uparrow$ \\
\midrule
RMM-FOSTER & \textbf{11.17} & 5.186 & 66.10 & 12.7 \\
MEMO & 170.60 & 3.231 & 60.94 & 18.8 \\
DER & 223.40 & 5.328 & 64.40 & 12.1 \\
TagFex & 128.18 & 15.362 & 73.32 & 4.8 \\
ACIL & \textbf{11.17} & \textbf{0.204} & 88.48 & \textbf{433.73} \\
DS-AL & \textbf{11.17} & 0.308 & \textbf{88.82} & 288.37 \\
\midrule
CIFNet (Ours) & \textbf{11.17} & 0.264 & 81.04 & 306.97 \\
\bottomrule
\end{tabular}
}
\end{table}

Tables~\ref{tab:cifar100-efficiency} and~\ref{tab:imagenet100-efficiency}, together with Figure~\ref{fig:bubble_green}, compare the computational cost of representative methods spanning the main CIL paradigms. We focus on CIFAR-100 and ImageNet-100, where differences in computational cost become most evident.

Among iterative methods, TagFex achieves the highest $A_K$ on CIFAR-100 ($69.30\%$), but at a substantially higher computational cost than the remaining approaches. It consumes $1.151$\,kWh and $15.362$\,kWh on CIFAR-100 and ImageNet-100 respectively. Its energy consumption is approximately $19\times$ and $58\times$ the energy expenditure of CIFNet, in CIFAR-100 and ImageNet-100 respectively. These large increases in energy consumption translate into relatively modest accuracy improvements (that disappear entirely on CORe50). Together, these results suggest that the benefits of continuous backbone optimisation become increasingly dataset dependent, whereas its computational cost remains consistently high. This trend is also reflected in Figure~\ref{fig:bubble_green}.  TagFex occupies the high-accuracy region but requires substantially longer training times and higher energy consumption. By contrast, frozen-backbone analytic methods consistently achieve much higher accuracy-per-energy ratios, illustrating a more favourable trade-off between predictive performance and computational cost.

Within the analytic family, CIFNet achieves the highest accuracy-per-energy ratio ($A_K$/kWh) on CIFAR-100 (Table~\ref{tab:cifar100-efficiency}), despite not obtaining the highest final accuracy. This efficiency stems from operating directly in the backbone embedding space without maintaining the high-dimensional projected representations required by ACIl and DS-AL. On ImageNet-100 (Table~\ref{tab:imagenet100-efficiency}), ACIL achieves the highest efficiency ratio, with CIFNet and DS-AL following closely. Importantly, all three analytic methods remain substantially more energy efficient iterative approaches. The remaining differences within the analytic family primarily reflect the computational cost of maintaining and inverting covariance matrices, whose size grows with the projection dimension.

\begin{figure*}
    \centering
    \includegraphics[width=0.8\linewidth]{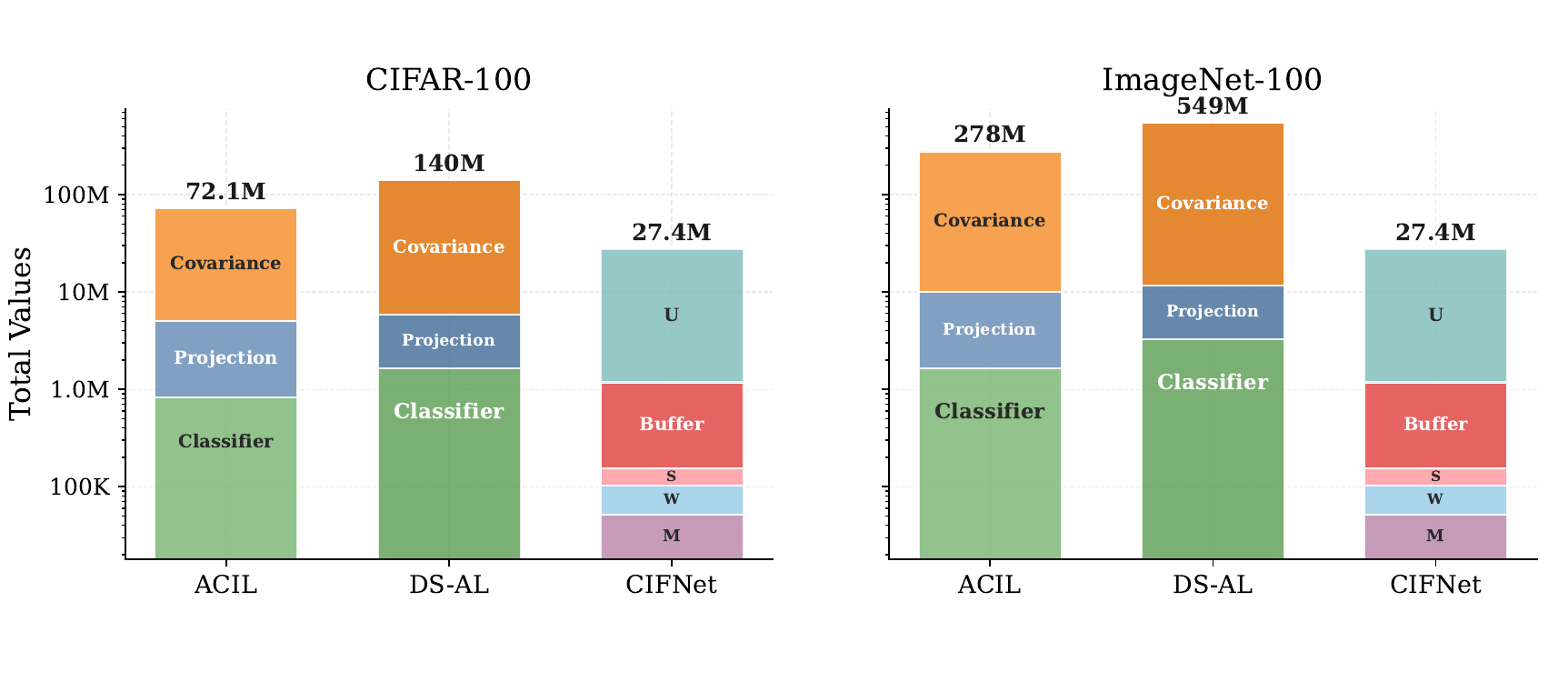}
    \caption{Classifier memory footprint of analytic CIL methods, measured in total number elements on logarithmic scale. Bars decomposed by component: ACIL and DS-AL maintain a fixed random projection matrix, a covariance matrix, and a classifier weight matrix in a high-dimensional projected space of width $B$; CIFNet maintains per-class RRLS sufficient statistics ($U$, $S$, $M$) and a compact calibration buffer directly in the backbone embedding space. CIFNet's footprint (27.4M elements) is identical across both benchmarks, reflecting its independence from auxiliary projection dimensionality.}
    \label{fig:memory}
\end{figure*}

Figure~\ref{fig:memory} extends the efficiency analysis by comparing the classifier memory footprint, measured as the total number of scalar elements to provide a precision-independent comparison. ACIL and DS-AL operate in a projected space of dimension $B \gg d$, causing their memory requirements to be dominated by $\mathcal{O}(B^2)$ covariance matrices. This results in 72.1M and 140M stored elements on CIFAR-100 ($B=8{,}129$), and 278M and 549M on ImageNet-100 ($B=16{,}384$), for ACIL and DS-AL respectively. CIFNet instead stores only the per-class RRLS sufficient statistics together with a compact calibration buffer directly in the backbone embedding space, requiring 27.4M elements regardless of the benchmark. This invariance follows directly from CIFNet scaling with the fixed backbone dimension $d$ and the number of classes $|\mathcal{C}|$, rather than with the projection dimension $B$, as detailed in Appendix~\ref{app:memory}. In a deployment scenario with limited resources, CIFNet requires between $2.5\times$ and $10.1\times$ fewer classifier memory than ACIL, and between $5\times$ and $20\times$ less than DS-AL. These reductions facilitate deployment on edge and embedded platforms, where memory budgets often represent a stricter constraint than raw computational throughput.

\subsection{Comparison Within the Analytic Learning Family}
\label{subsec:analytic_family}

Having established the overall accuracy, trajectory, and efficiency trade-offs, we now examine the methodological differences within the analytic family, where CIFNet, ACIL, and DS-AL share the same frozen-backbone paradigm but make different design choices.

On CIFAR-100 and ImageNet-100, ACIL and DS-AL consistently achieve greater $A_K$ than CIFNet, typically by 7-8 percentage points. This advantage is largely attributable to their use of a high-dimensional projected feature space ($B \gg d$), which increases the linear separability of class representations before solving the analytic classification problem. CIFNet instead performs classification directly in the backbone embedding space, deliberately sacrificing some representational capacity in exchange for substantially lower memory usage and computational cost. The three methods therefore occupy different operating points within the analytic CIL design space rather than representing strictly superior or inferior alternatives.

As shown in previous subsections, this trade-off extends beyond final accuracy. CIFNet provides smoother learning trajectories and consistently higher computational efficiency, while ACIL and DS-AL prioritise end-point performance through larger projected representations. On CORe50, this balance shifts further in favour of CIFNet, which achieves the highest average accuracy among analytic methods. A plausible explanation is that the stationary embedding space and calibrated updates reduce the impact of the strong inter-class similarity characteristic of CORe50, whereas methods relying on increasingly complex decision boundaries appear more sensitive to this setting.

To further analyse these behavioural differences, we evaluated all analytic methods under a substantial longer incremental sequence comprising 50 tasks. Standard CIL benchmarks involve at most 20 sequential updates, whereas practical deployments may require considerably longer learning horizons. This experiment therefore examines whether the trajectory patterns observed previously persist under more demanding continual learning conditions.

\begin{figure}[ht]
    \centering
    \includegraphics[width=0.85\linewidth]{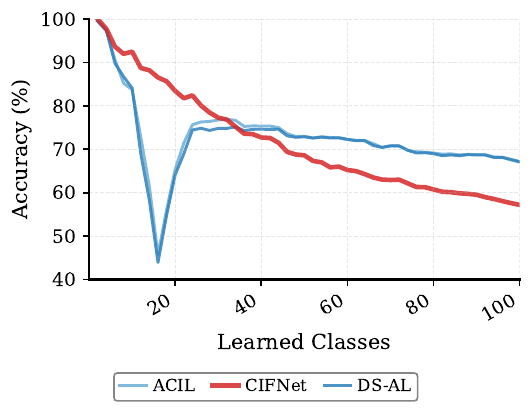}
    \caption{Incremental accuracy ($A_k$) of analytic CIL methods on CIFAR-100 under the 50-task setting (2 classes per task). CIFNet maintains a monotonically decreasing trajectory throughout the sequence, whereas ACIL and DS-AL exhibit severe accuracy collapses at intermediate tasks before recovering as the full class structure is absorbed.}
    \label{fig:long_tail}
\end{figure}

Figure~\ref{fig:long_tail} highlights a behavioural difference that is not apparent from the final accuracy alone. Although ACIL and DS-AL achieve higher $A_K$, both methods exhibit abrupt performance collapses at several intermediate tasks, losing approximately 20 percentage points after a single incremental update before recovering as additional classes are incorporated. These oscillations suggest that expanding the classifier without explicitly calibrating newly introduced decision boundaries can temporarily destabilise previously acquired knowledge.

By contrast, CIFNet task maintains the same smooth, monotonically decreasing trajectory observed in the standard benchmarks, even across 50 sequential tasks. Although its final accuracy remains lower ($57.23\%$ on CIFAR-100), the gap in average accuracy is considerably smaller ($71.60\%$ versus $72.73\%$ and $72.14\%$ on CIFAR-100). For deployment scenarios involving continuous operation, these trajectory characteristics may be more relevant than the final task performance alone. Edge devices, robotic systems, and online learning applications require reliable predictions throughout the incremental sequence rather than only after convergence. From this perspective, CIFNet offers a complementary operating point within the analytic family by prioritising trajectory stability and computational efficiency while maintaining competitive average performance.

\subsection{Ablation Study}
\label{subsec:ablation}

To validate the key design choices of CIFNet, we conduct ablation studies on CIFAR-100 under 5- and 10-class increments (20 and 10 tasks, respectively). The analysis covers three aspects: the contribution of CIFNet-specific components, the effect of classifier choice under a frozen representation, and the sensitivity to buffer size and sampling strategy.

\paragraph{Effect of CIFNet Components.}

\begin{table}[ht]
    \caption{Impact of CIFNet components on accuracy ($\bar{A}$ and $A_K$) using the CIFAR-100 dataset with 5 and 10 classes per task.}
    \label{tab:ablation}
    \centering
    \small
    \resizebox{\columnwidth}{!}{
    \begin{tabular}{lrrrr}
        \toprule
        \multirow{2}{*}{} & \multicolumn{2}{c}{5 cls/task} & \multicolumn{2}{c}{10 cls/task} \\
        & $\bar{A} \uparrow$ & $A_K \uparrow$ & $\bar{A} \uparrow$ & $A_K \uparrow$ \\
        \midrule
        Without Calibration Buffer & 42.18 & 19.42 & 50.36 & 24.74 \\
        Without Oversampling     & 61.74 & 36.62 & 61.63 & 37.55 \\
        Full Method              & 73.56 & 59.26 & 72.93 & 59.57 \\
        \bottomrule
        \end{tabular}
    }
\end{table}

\begin{figure}[ht]
    \centering
    \includegraphics[width=\linewidth]{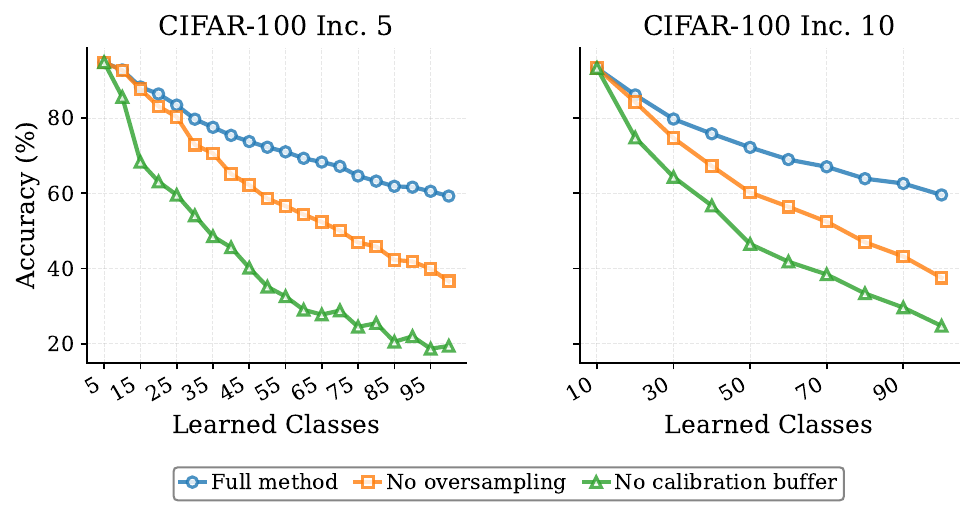}
    \caption{Incremental learning performance ($A_k$) of CIFNet on CIFAR-100 under different class-increment settings, comparing variants with key components removed.}
    \label{fig:ablation_curve}
\end{figure}

Table~\ref{tab:ablation} and Figure~\ref{fig:ablation_curve} isolate the contribution of the calibration buffer and the density-aware oversampling strategy. Removing either component leads to clear performance degradation. Disabling the calibration buffer produces the most severe drop in both $\bar{A}$ and $A_K$ across both increment settings, confirming that providing new neurons with negative evidence from past classes is the single most important component of CIFNet's design. Disabling oversampling alone produces a more moderate degradation, concentrated in $A_K$, consistent with the interpretation that density imbalance primarily affects the calibration quality at later tasks when the gap between current-task sample count and buffer coverage is largest. The full CIFNet configuration consistently achieves the best results across both settings.

\paragraph{Classifier Choice under Frozen Representation.}

\begin{figure}[ht]
    \centering
    \includegraphics[width=\linewidth]{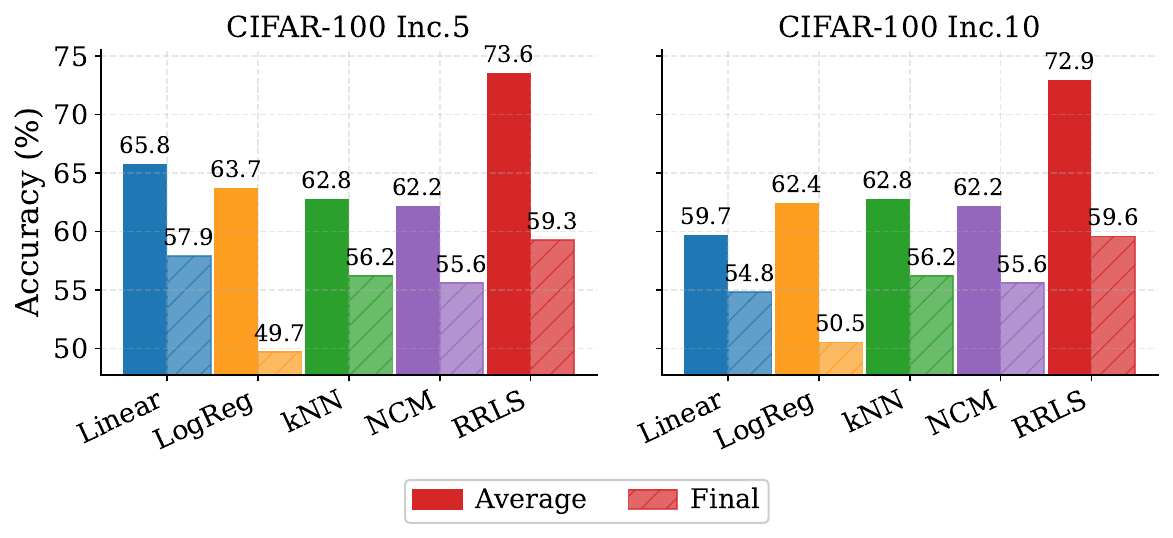}
    \caption{Incremental learning performance ($A_k$) on CIFAR-100 using a frozen ResNet-18 backbone with different classifier heads: linear classifier trained via SGD, multinomial logistic regression, k-Nearest Neighbours, nearest-class-mean classifier and the RRLS analytic classifier.}
    \label{fig:classifier_ablation}
\end{figure}

To isolate the effect of the analytic classifier, we compare CIFNet against alternative classifier heads trained on the same frozen feature extractor under identical class-increment protocols: a linear classifier trained with SGD, multinomial logistic regression, k-Nearest Neighbours (kNN), and a nearest-class-mean (NCM) classifier. As shown in Figure~\ref{fig:classifier_ablation}, linear and logistic regression classifiers achieve the lowest performance, with limited robustness across incremental steps and a pronounced gap between $\bar{A}$ and $A_K$ indicating instability in incremental adaptation. The kNN and NCM classifiers improve both $\bar{A}$ and $A_K$ due to their non-parametric nature, but require storing the full dataset (or the prototypes instead). The RRLS classifier consistently achieves the best trade-off between accuracy and memory footprint, outperforming all alternatives in both average and final accuracy across both incremental settings. The performance of CIFNet is not only a consequence of the frozen backbone, but depends fundamentally on the specific analytic classifier mechanism.

\paragraph{Buffer Size and Sampling Strategy.}

\begin{figure}[t]
    \centering
    \includegraphics[width=0.75\linewidth]{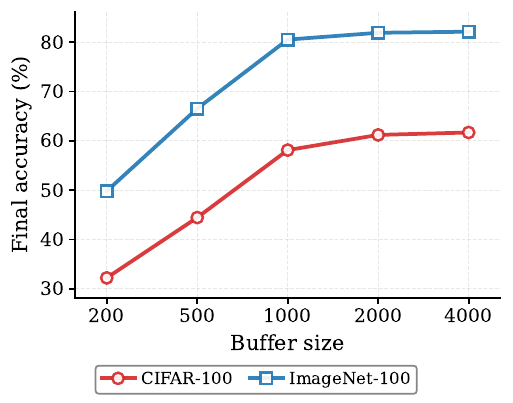}
    \caption{Effect of buffer size on CIFAR-100 and ImageNet-100. Performance is reported in terms of final accuracy ($A_K$).}
    \label{fig:buffer_ablation}
\end{figure}

Figure~\ref{fig:buffer_ablation} evaluates buffer capacities of $\{200, 500, 1{,}000, 2{,}000, 4{,}000\}$ samples on CIFAR-100 and ImageNet-100. The results show a clear saturation effect, where performance improves significantly from 200 to 1000 samples, then diminishes beyond 2,000. On CIFAR-100, 2,000 samples provide a strong balance between accuracy and efficiency; on ImageNet-100, performance stabilises beyond 1,000 samples, indicating that larger buffers do not yield meaningful gains in the higher-quality representation space available on that benchmark.

\begin{table}[t]
\centering
\caption{Effect of buffer sampling strategy on final accuracy ($A_K$, \%) on CIFAR-100 with 5 classes per task. Best result in \textbf{bold}.}
\label{tab:sampling}
\small
\begin{tabular}{lc}
    \toprule
    Sampling Strategy & $A_K$ (\%) \\
    \midrule
    \textbf{Random} & \textbf{59.26} \\
    $k$-Means       & 58.25 \\
    Herding         & 58.07 \\
    Centroid        & 57.26 \\
    \bottomrule
\end{tabular}
\end{table}

Finally, Table~\ref{tab:sampling} reports the effect of buffer sampling strategy on final accuracy ($A_K$) on CIFAR-100 with 5 classes per task. Four strategies are evaluated: random sampling, herding, centroid-based selection, and $k$-means clustering. Random sampling achieves the highest final accuracy (59.26\%), outperforming herding (58.07\%), $k$-means (58.25\%), and centroid selection (57.26\%). The differences are modest, spanning a range of approximately two percentage points, which suggests that the calibration mechanism is not highly sensitive to the choice of sampling strategy. The advantage of random sampling is consistent with the theoretical motivation in Section~\ref{subsec:buffer}, selection-based approaches concentrate the buffer on prototypical embeddings, underrepresenting distributional tails and introducing a systematic skew into the RRLS covariance estimation that random sampling avoids.

\section{Conclusions and Future Work}
\label{sec:conclusions}

This work introduces CIFNet, a frugal and efficient analytic framework for class-incremental learning that shifts continual adaptation from iterative gradient-based optimization to closed-form estimation. By freezing feature representations and updating the classification layer via Regularised Recursive Least-Squares (RRLS), CIFNet eliminates backpropagation, learning-rate scheduling, and stochastic convergence issues. The proposed latent calibration buffer and density-aware oversampling effectively mitigate the structural initialisation bias inherent to output neuron expansion, establishing balanced decision boundaries across cumulative label spaces.

Empirical evaluations across CIFAR-100, ImageNet-100, CORe50, and an extended 50-task sequence demonstrate that competitive CIL accuracy does not strictly require continuous backbone re-training. Crucially, CIFNet exhibits smooth, strictly monotonic accuracy trajectories, avoiding the transient performance collapses observed in uncalibrated analytic methods. By reducing incremental updates to closed-form moment accumulation, CIFNet achieves up to a 20$\times$ energy reduction relative to gradient-trained baselines, reinforcing analytic learning as a mathematically rigorous and efficient paradigm for lifelong systems.

Beyond immediate compute savings, this perspective aligns with the broader goals of long-horizon continual learning. As systems operate over extended timelines and across numerous task iterations, replacing backpropagation with analytic updates significantly reduces operational energy footprints without compromising predictive fidelity.

Despite its advantages, CIFNet has specific limitations. It assumes that the frozen representations remain sufficiently informative throughout the incremental sequence; under severe domain shifts, classifier-level adaptation alone may prove insufficient. Furthermore, the memory and computational requirements of RRLS sufficient statistics scale with feature dimensionality, which may constrain deployment alongside very large vision foundation models.

Future work will explore two complementary directions. First, integrating parameter-efficient adaptation mechanisms (e.g., low-rank adaptation or lightweight adapters) could introduce controlled backbone plasticity while preserving analytic efficiency. Second, the additive structure of RRLS sufficient statistics naturally extends to federated class-incremental learning, enabling collaborative multi-client model updates without exchanging raw data or private gradients.

\section*{Acknowledgements}

Work funded by Project PID2023-147404OB-I00 (MICIU / AEI / 10.13039 / 501100011033; ERDF/EU; ESF+/EU), Horizon Europe (GA 101070381), the Ministry for Digital Transformation and Civil Service and Next-GenerationEU/PRTR (TSI-100925-2023-1) and by the Competitive Reference Groups of the Xunta de Galicia (grant ED431C 2026/38).  CITIC, as a member of the CIGUS Network, receives subsidies from the ``Xunta de Galicia" and from the ERDF Operational Programme Galicia 2021-2027 (Grant ED431G 2023/01).

\bibliographystyle{ieeetr}

\bibliography{bibliography}

@inproceedings{schlimmer1986case,
  title={A case study of incremental concept induction},
  author={Schlimmer, Jeffrey C and Fisher, Douglas},
  booktitle={Proceedings of the Fifth AAAI National Conference on Artificial Intelligence},
  pages={496--501},
  year={1986}
}

@misc{Parisi2019,
   author = {German I. Parisi and Ronald Kemker and Jose L. Part and Christopher Kanan and Stefan Wermter},
   doi = {10.1016/j.neunet.2019.01.012},
   issn = {18792782},
   journal = {Neural Networks},
   month = {5},
   pages = {54-71},
   pmid = {30780045},
   publisher = {Elsevier Ltd},
   title = {Continual lifelong learning with neural networks: A review},
   volume = {113},
   year = {2019},
}

@article{Rebuffi2017,
   author = {Sylvestre Alvise Rebuffi and Alexander Kolesnikov and Georg Sperl and Christoph H. Lampert},
   doi = {10.1109/CVPR.2017.587},
   isbn = {9781538604571},
   journal = {Proceedings - 30th IEEE Conference on Computer Vision and Pattern Recognition, CVPR 2017},
   month = {11},
   pages = {5533-5542},
   publisher = {Institute of Electrical and Electronics Engineers Inc.},
   title = {iCaRL: Incremental classifier and representation learning},
   volume = {2017-January},
   year = {2017},
}

@article{McCloskey1989,
   author = {Michael McCloskey and Neal J. Cohen},
   doi = {10.1016/S0079-7421(08)60536-8},
   issn = {0079-7421},
   issue = {C},
   journal = {Psychology of Learning and Motivation - Advances in Research and Theory},
   month = {1},
   pages = {109-165},
   publisher = {Academic Press},
   title = {Catastrophic Interference in Connectionist Networks: The Sequential Learning Problem},
   volume = {24},
   year = {1989},
}

@article{barry2023neural,
  title={A neural network account of memory replay and knowledge consolidation},
  author={Barry, Daniel N and Love, Bradley C},
  journal={Cerebral Cortex},
  volume={33},
  number={1},
  pages={83--95},
  year={2023},
  publisher={Oxford University Press}
}

@article{rusu2016progressive,
  title={Progressive neural networks},
  author={Rusu, Andrei A and Rabinowitz, Neil C and Desjardins, Guillaume and Soyer, Hubert and Kirkpatrick, James and Kavukcuoglu, Koray and Pascanu, Razvan and Hadsell, Raia},
  journal={arXiv preprint arXiv:1606.04671},
  year={2016}
}

@inproceedings{mallya2018packnet,
  title={Packnet: Adding multiple tasks to a single network by iterative pruning},
  author={Mallya, Arun and Lazebnik, Svetlana},
  booktitle={Proceedings of the IEEE conference on Computer Vision and Pattern Recognition},
  pages={7765--7773},
  year={2018}
}

@article{krizhevsky2009learning,
  title={Learning multiple layers of features from tiny images},
  author={Krizhevsky, Alex and Hinton, Geoffrey and others},
  year={2009},
  publisher={Toronto, ON, Canada}
}

@article{le2015tiny,
  title={Tiny imagenet visual recognition challenge},
  author={Le, Ya and Yang, Xuan},
  journal={CS 231N},
  volume={7},
  number={7},
  pages={3},
  year={2015}
}

@inproceedings{fontenla2021rolann,
author = {Fontenla-Romero, Oscar and Guijarro-Berdi\~{n}as, Bertha and P\'{e}rez-S\'{a}nchez, Beatriz},
title = {Regularized One-Layer Neural Networks for Distributed and Incremental Environments},
year = {2021},
isbn = {978-3-030-85098-2},
publisher = {Springer-Verlag},
address = {Berlin, Heidelberg},
doi = {10.1007/978-3-030-85099-9\_28},
booktitle = {Advances in Computational Intelligence: 16th International Work-Conference on Artificial Neural Networks, IWANN 2021, Virtual Event, June 16–18, 2021, Proceedings, Part II},
pages = {343–355},
numpages = {13}
}

@article{zhou2024cil_survey,
    author = {Zhou, Da-Wei and Wang, Qi-Wei and Qi, Zhi-Hong and Ye, Han-Jia and Zhan, De-Chuan and Liu, Ziwei},
    title = {Class-Incremental Learning: A Survey},
    journal={IEEE Transactions on Pattern Analysis and Machine Intelligence},
    volume={46},
    number={12},
    pages={9851--9873},
    year = {2024}
 }

@article{buzzega2020dark,
  title={Dark experience for general continual learning: a strong, simple baseline},
  author={Buzzega, Pietro and Boschini, Matteo and Porrello, Angelo and Abati, Davide and Calderara, Simone},
  journal={Advances in neural information processing systems},
  volume={33},
  pages={15920--15930},
  year={2020}
}

@inproceedings{he2016deep,
  title={Deep residual learning for image recognition},
  author={He, Kaiming and Zhang, Xiangyu and Ren, Shaoqing and Sun, Jian},
  booktitle={Proceedings of the IEEE conference on computer vision and pattern recognition},
  pages={770--778},
  year={2016}
}

@inproceedings{wang2022foster,
  title={Foster: Feature boosting and compression for class-incremental learning},
  author={Wang, Fu-Yun and Zhou, Da-Wei and Ye, Han-Jia and Zhan, De-Chuan},
  booktitle={European conference on computer vision},
  pages={398--414},
  year={2022},
  organization={Springer}
}

@article{zhou2022model,
  title={A model or 603 exemplars: Towards memory-efficient class-incremental learning},
  author={Zhou, Da-Wei and Wang, Qi-Wei and Ye, Han-Jia and Zhan, De-Chuan},
  journal={arXiv preprint arXiv:2205.13218},
  year={2022}
}

@inproceedings{douillard2020podnet,
  title={Podnet: Pooled outputs distillation for small-tasks incremental learning},
  author={Douillard, Arthur and Cord, Matthieu and Ollion, Charles and Robert, Thomas and Valle, Eduardo},
  booktitle={Computer vision--ECCV 2020: 16th European conference, Glasgow, UK, August 23--28, 2020, proceedings, part XX 16},
  pages={86--102},
  year={2020},
  organization={Springer}
}

@inproceedings{wu2019large,
  title={Large scale incremental learning},
  author={Wu, Yue and Chen, Yinpeng and Wang, Lijuan and Ye, Yuancheng and Liu, Zicheng and Guo, Yandong and Fu, Yun},
  booktitle={Proceedings of the IEEE/CVF conference on computer vision and pattern recognition},
  pages={374--382},
  year={2019}
}

@inproceedings{jung2023generating,
  title={Generating instance-level prompts for rehearsal-free continual learning},
  author={Jung, Dahuin and Han, Dongyoon and Bang, Jihwan and Song, Hwanjun},
  booktitle={Proceedings of the IEEE/CVF International Conference on Computer Vision},
  pages={11847--11857},
  year={2023}
}

@article{zhou2025revisiting,
  title={Revisiting class-incremental learning with pre-trained models: Generalizability and adaptivity are all you need},
  author={Zhou, Da-Wei and Cai, Zi-Wen and Ye, Han-Jia and Zhan, De-Chuan and Liu, Ziwei},
  journal={International Journal of Computer Vision},
  volume={133},
  number={3},
  pages={1012--1032},
  year={2025},
  publisher={Springer}
}

@inproceedings{chen2025reducing,
  title={Reducing Class-wise Confusion for Incremental Learning with Disentangled Manifolds},
  author={Chen, Huitong and Wang, Yu and Fan, Yan and Jiang, Guosong and Hu, Qinghua},
  booktitle={Proceedings of the Computer Vision and Pattern Recognition Conference},
  pages={10121--10130},
  year={2025}
}

@inproceedings{zheng2025task,
  title={Task-Agnostic Guided Feature Expansion for Class-Incremental Learning},
  author={Zheng, Bowen and Zhou, Da-Wei and Ye, Han-Jia and Zhan, De-Chuan},
  booktitle={Proceedings of the Computer Vision and Pattern Recognition Conference},
  pages={10099--10109},
  year={2025}
}

@inproceedings{wang2022learning,
  title={Learning to prompt for continual learning},
  author={Wang, Zifeng and Zhang, Zizhao and Lee, Chen-Yu and Zhang, Han and Sun, Ruoxi and Ren, Xiaoqi and Su, Guolong and Perot, Vincent and Dy, Jennifer and Pfister, Tomas},
  booktitle={Proceedings of the IEEE/CVF Conference on Computer Vision and Pattern Recognition},
  pages={139--149},
  year={2022}
}

@InProceedings{Smith_2023_CVPR,
    author    = {Smith, James Seale and Karlinsky, Leonid and Gutta, Vyshnavi and Cascante-Bonilla, Paola and Kim, Donghyun and Arbelle, Assaf and Panda, Rameswar and Feris, Rogerio and Kira, Zsolt},
    title     = {CODA-Prompt: COntinual Decomposed Attention-Based Prompting for Rehearsal-Free Continual Learning},
    booktitle = {Proceedings of the IEEE/CVF Conference on Computer Vision and Pattern Recognition (CVPR)},
    month     = {June},
    year      = {2023},
    pages     = {11909-11919}
}

@inproceedings{gao2025maintaining,
  title={Maintaining fairness in logit-based knowledge distillation for class-incremental learning},
  author={Gao, Zijian and Han, Shanhao and Zhang, Xingxing and Xu, Kele and Zhou, Dulan and Mao, Xinjun and Dou, Yong and Wang, Huaimin},
  booktitle={Proceedings of the AAAI Conference on Artificial Intelligence},
  volume={39},
  number={16},
  pages={16763--16771},
  year={2025}
}

@inproceedings{szatkowski2024adapt,
  title={Adapt your teacher: Improving knowledge distillation for exemplar-free continual learning},
  author={Szatkowski, Filip and Pyla, Mateusz and Przewi{\k{e}}{\'z}likowski, Marcin and Cygert, Sebastian and Twardowski, Bart{\l}omiej and Trzci{\'n}ski, Tomasz},
  booktitle={Proceedings of the IEEE/CVF Winter Conference on Applications of Computer Vision},
  pages={1977--1987},
  year={2024}
}

@article{li2017learning,
  title={Learning without forgetting},
  author={Li, Zhizhong and Hoiem, Derek},
  journal={IEEE transactions on pattern analysis and machine intelligence},
  volume={40},
  number={12},
  pages={2935--2947},
  year={2017},
  publisher={IEEE}
}

@article{kirkpatrick2017overcoming,
  title={Overcoming catastrophic forgetting in neural networks},
  author={Kirkpatrick, James and Pascanu, Razvan and Rabinowitz, Neil and Veness, Joel and Desjardins, Guillaume and Rusu, Andrei A and Milan, Kieran and Quan, John and Ramalho, Tiago and Grabska-Barwinska, Agnieszka and others},
  journal={Proceedings of the national academy of sciences},
  volume={114},
  number={13},
  pages={3521--3526},
  year={2017},
  publisher={National Academy of Sciences}
}

@article{balasubramanian2024exacfs,
  title={EXACFS--A CIL Method to mitigate Catastrophic Forgetting},
  author={Balasubramanian, S and Subramaniam, M Sai and Talasu, Sai Sriram and Sai, Manepalli Pranav Phanindra and Mukkamala, Ravi and Gera, Darshan and others},
  journal={arXiv preprint arXiv:2410.23751},
  year={2024}
}

@article{zhuang2022acil,
  title={ACIL: Analytic class-incremental learning with absolute memorization and privacy protection},
  author={Zhuang, Huiping and Weng, Zhenyu and Wei, Hongxin and Xie, Renchunzi and Toh, Kar-Ann and Lin, Zhiping},
  journal={Advances in Neural Information Processing Systems},
  volume={35},
  pages={11602--11614},
  year={2022}
}

@inproceedings{zhuang2024ds,
  title={DS-AL: A dual-stream analytic learning for exemplar-free class-incremental learning},
  author={Zhuang, Huiping and He, Run and Tong, Kai and Zeng, Ziqian and Chen, Cen and Lin, Zhiping},
  booktitle={Proceedings of the AAAI Conference on Artificial Intelligence},
  volume={38},
  number={15},
  pages={17237--17244},
  year={2024}
}

@article{li2023CRNet,
  title={{CRNet}: A Fast Continual Learning Framework With Random Theory}, 
  author={Li, Depeng and Zeng, Zhigang},
  journal={IEEE Transactions on Pattern Analysis and Machine Intelligence}, 
  volume={45},
  number={9},
  pages={10731--10744},
  year={2023},
  publisher={IEEE}
}

@inproceedings{zhou2024expandable,
  title={Expandable subspace ensemble for pre-trained model-based class-incremental learning},
  author={Zhou, Da-Wei and Sun, Hai-Long and Ye, Han-Jia and Zhan, De-Chuan},
  booktitle={Proceedings of the IEEE/CVF Conference on Computer Vision and Pattern Recognition},
  pages={23554--23564},
  year={2024}
}

@inproceedings{zhang2023slca,
  title={Slca: Slow learner with classifier alignment for continual learning on a pre-trained model},
  author={Zhang, Gengwei and Wang, Liyuan and Kang, Guoliang and Chen, Ling and Wei, Yunchao},
  booktitle={Proceedings of the IEEE/CVF International Conference on Computer Vision},
  pages={19148--19158},
  year={2023}
}

@article{wang2023hierarchical,
  title={Hierarchical decomposition of prompt-based continual learning: Rethinking obscured sub-optimality},
  author={Wang, Liyuan and Xie, Jingyi and Zhang, Xingxing and Huang, Mingyi and Su, Hang and Zhu, Jun},
  journal={Advances in Neural Information Processing Systems},
  volume={36},
  pages={69054--69076},
  year={2023}
}

@article{le2024mixture,
  title={Mixture of experts meets prompt-based continual learning},
  author={Le, Minh and Nguyen, An and Nguyen, Huy and Nguyen, Trang and Pham, Trang and Van Ngo, Linh and Ho, Nhat},
  journal={Advances in Neural Information Processing Systems},
  volume={37},
  pages={119025--119062},
  year={2024}
}

@article{brand2006fast,
title = {Fast low-rank modifications of the thin singular value decomposition},
journal = {Linear Algebra and its Applications},
volume = {415},
number = {1},
pages = {20-30},
year = {2006},
note = {Special Issue on Large Scale Linear and Nonlinear Eigenvalue Problems},
issn = {0024-3795},
doi = {https://doi.org/10.1016/j.laa.2005.07.021},
url = {https://www.sciencedirect.com/science/article/pii/S0024379505003812},
author = {Matthew Brand},
}

@inproceedings{lomonaco2017core50,
  title={Core50: a new dataset and benchmark for continuous object recognition},
  author={Lomonaco, Vincenzo and Maltoni, Davide},
  booktitle={Conference on robot learning},
  pages={17--26},
  year={2017},
  organization={PMLR}
}

@inproceedings{strubell2019energy,
    title = "Energy and Policy Considerations for Deep Learning in {NLP}",
    author = "Strubell, Emma  and
      Ganesh, Ananya  and
      McCallum, Andrew",
    editor = "Korhonen, Anna  and
      Traum, David  and
      M{\`a}rquez, Llu{\'i}s",
    booktitle = "Proceedings of the 57th Annual Meeting of the Association for Computational Linguistics",
    month = jul,
    year = "2019",
    address = "Florence, Italy",
    publisher = "Association for Computational Linguistics",
    url = "https://aclanthology.org/P19-1355/",
    doi = "10.18653/v1/P19-1355",
    pages = "3645--3650",
}

@software{codecarbon,
  author       = {Benoit Courty and
                  Victor Schmidt and
                  Sasha Luccioni and
                  Goyal-Kamal and
                  MarionCoutarel and
                  Boris Feld and
                  Jérémy Lecourt and
                  LiamConnell and
                  Amine Saboni and
                  Inimaz and
                  supatomic and
                  Mathilde Léval and
                  Luis Blanche and
                  Alexis Cruveiller and
                  ouminasara and
                  Franklin Zhao and
                  Aditya Joshi and
                  Alexis Bogroff and
                  Hugues de Lavoreille and
                  Niko Laskaris and
                  Edoardo Abati and
                  Douglas Blank and
                  Ziyao Wang and
                  Armin Catovic and
                  Marc Alencon and
                  Michał Stęchły and
                  Christian Bauer and
                  Lucas Otávio N. de Araújo and
                  JPW and
                  MinervaBooks},
  title        = {mlco2/codecarbon: v2.4.1},
  month        = may,
  year         = 2024,
  publisher    = {Zenodo},
  version      = {v2.4.1},
  doi          = {10.5281/zenodo.11171501},
  url          = {https://doi.org/10.5281/zenodo.11171501}
}

\appendix
\section{Proof of the RRLS Closed-Form Solution}
\label{app:rrls-proof}

The closed-form solution in Equation~\eqref{eq:weights_rrls} is the unique minimiser of the regularised weighted least-squares objective.

\begin{theorem}[Global Optimality of the RRLS Solution]

Let $X$, $F$, $\hat{y}$, and $\lambda>0$ be defined as in Section~\ref{subsec:rrls}. Then the weight vector

\begin{equation}
    w=U(S^2+\lambda I)^{-1}U^\top m
\end{equation}

\noindent is the unique minimiser of

\begin{equation}
    \min_w \; \|H^\top w-F\hat y\|_2^2 + \lambda\|w\|_2^2.
    \label{eq:objective_rrls}
\end{equation}

\end{theorem}

\begin{proof}

Setting $\nabla_w\mathcal L=0$ yields the normal equations

\begin{equation}
(HH^\top+\lambda I)w=m,
\end{equation}

\noindent where $m$ is given by Equation~\eqref{eq:moment_rrls}. Since $\lambda>0$, the matrix $HH^\top+\lambda I$ is strictly positive definite, guaranteeing both existence and uniqueness of the solution.

Using the reduced singular value decomposition $H=USV^\top$, we obtain

\begin{equation}
HH^\top=US^2U^\top.
\end{equation}

Substituting this expression into the normal equations gives

\begin{equation}
(US^2U^\top+\lambda I)w=m.
\end{equation}

Because the minimiser belongs to the column space of $H$, we have $w=UU^\top w$. Premultiplying by $U^\top$ yields

\begin{equation}
(S^2+\lambda I)(U^\top w)=U^\top m.
\end{equation}

Finally,

\begin{equation}
U^\top w=(S^2+\lambda I)^{-1}U^\top m,
\end{equation}

\noindent and multiplying both sides by $U$ recovers

\begin{equation}
w=
U(S^2+\lambda I)^{-1}U^\top m,
\end{equation}

\noindent which is precisely Equation~\eqref{eq:weights_rrls}.

\end{proof}

\section{Memory Complexity of Analytic CIL Methods}
\label{app:memory}

This appendix derives the persistent memory footprint of ACIL, DS-AL, and CIFNet as reported in Section~\ref{sec:evaluation}. We focus exclusively on classifier state stored across incremental tasks, excluding backbone parameters, which are identical across all three methods. All element counts correspond to scalar float32 values; byte sizes are computed as $\text{elements} \times 4$.

\subsection*{Notation}

Let $D$ denote the backbone feature dimension ($d=512$ for ResNet-18), $C$ the total number of classes, and $B$ the random projection width used in ACIL and DS-AL. CIFNet operates directly in the native embedding space without projection. The effective rank of the RRLS factorisation is $r = \min(d+1, N_k)$, where $N_k$ is the total number of samples observed up to the current task $k$; in practice, $N_k \gg d+1 \implies = d+1$.

\subsection*{ACIL}

ACIL \cite{zhuang2022acil} maintains three persistent components: a projection matrix $\mathbf{P} \in \mathbb{R}^{B \times d}$, a covariance matrix $\mathbf{R} \in \mathbb{R}^{B \times B}$, and a classifier weight matrix $\mathbf{W} \in \mathbb{R}^{C \times B}$. The total element count is:

\begin{equation}
    \mathcal{M}_{\text{ACIL}} = Bd + B^2 + CB
\end{equation}

\noindent which is $\mathcal{O}(B^2)$ when $d$ and $C$ are fixed, since the covariance term dominates for all practical values of $B$.

\subsection*{DS-AL}

DS-AL \cite{zhuang2024ds} introduces a second parallel analytic stream with the same projection width, effectively duplicating the covariance and classifier components:

\begin{equation}
    \mathcal{M}_{\text{DS-AL}} = Bd + 2B^2 + 2CB
\end{equation}

\noindent preserving the $\mathcal{O}(B^2)$ asymptotic scaling with a constant factor of approximately two relative to ACIL.

\subsection*{CIFNet}

CIFNet avoids feature projection entirely. Its persistent state consists of a calibration buffer of $B_{\text{cal}}$ embeddings in $\mathbb{R}^d$, and per-class sufficient statistics: a moment vector $M_c \in \mathbb{R}^{d+1}$, a left singular factor $U_c \in \mathbb{R}^{(d+1)\times r}$, singular values $S_c \in \mathbb{R}^r$, and a weight vector $w_c \in \mathbb{R}^{d+1}$. Aggregating over all $C$ classes:

\begin{equation}
    \mathcal{M}_{\text{CIFNet}} = B_{\text{cal}}\,d + C\bigl(2(d+1) + (d+1)r + r\bigr),
\end{equation}

\noindent which reduces to $\mathcal{O}(B_{\text{cal}}\,d + C\,d^2)$ when $r = d+1$. For fixed $d$ and $C$, this is a constant independent of the task horizon and of any projection hyperparameter.

\subsection*{Numerical Results and Scaling}

\begin{table*}
\centering
\caption{Detailed component breakdown of persistent classifier memory. Element counts (Elements) are in millions (M); memory sizes (Size) are in megabytes (MB).}
\label{tab:memory_breakdown}
\small
\begin{tabular}{llrrrr}
    \toprule
     & & \multicolumn{2}{c}{CIFAR-100 ($B = 8{,}192$)} & \multicolumn{2}{c}{ImageNet-100 ($B = 16{,}384$)} \\
    \cmidrule(lr){3-4} \cmidrule(lr){5-6}
    Method & Component & Elements & Size & Elements & Size \\
    \midrule
    \textbf{ACIL} & Projection ($\mathbf{P}$) & 4.2 M & 16.8 MB & 8.4 M & 33.6 MB \\
     & Covariance ($\mathbf{R}$) & 67.1 M & 268.4 MB & 268.4 M & 1,073.7 MB \\
     & Classifier ($\mathbf{W}$) & 0.8 M & 3.3 MB & 1.6 M & 6.6 MB \\
     \cmidrule(lr){2-6}
     & \textbf{Total} & \textbf{72.1 M} & \textbf{288.5 MB} & \textbf{278.4 M} & \textbf{1,113.9 MB} \\
    \midrule
    \textbf{DS-AL} & Projection ($\mathbf{P}$) & 4.2 M & 16.8 MB & 8.4 M & 33.6 MB \\
     & Covariance ($2\mathbf{R}$) & 134.2 M & 536.9 MB & 536.9 M & 2,147.5 MB \\
     & Classifier ($2\mathbf{W}$) & 1.6 M & 6.6 MB & 3.3 M & 13.1 MB \\
     \cmidrule(lr){2-6}
     & \textbf{Total} & \textbf{140.0 M} & \textbf{560.2 MB} & \textbf{548.6 M} & \textbf{2,194.5 MB} \\
    \midrule
    \textbf{CIFNet} & Buffer & 1.0 M & 4.1 MB & 1.0 M & 4.1 MB \\
     & Moments ($M$) & 0.05 M & 0.2 MB & 0.05 M & 0.2 MB \\
     & SVD factors ($U + S$) & 26.4 M & 105.5 MB & 26.4 M & 105.5 MB \\
     & Weights ($W$) & 0.05 M & 0.2 MB & 0.05 M & 0.2 MB \\
     \cmidrule(lr){2-6}
     & \textbf{Total} & \textbf{27.5 M} & \textbf{109.9 MB} & \textbf{27.5 M} & \textbf{109.9 MB} \\
    \bottomrule
\end{tabular}
\end{table*}

\begin{table}
\centering
\caption{Total memory footprint and asymptotic complexity comparison.}
\label{tab:memory_scaling}
\small
\resizebox{\columnwidth}{!}{
\begin{tabular}{lccl}
    \toprule
    Method & CIFAR-100 & ImageNet-100 & Complexity \\
    \midrule
    ACIL & 288.5 MB & 1,113.9 MB & $\mathcal{O}(B^2)$ \\
    DS-AL & 560.2 MB & 2,194.5 MB & $\mathcal{O}(B^2)$ \\ 
    \midrule
    \textbf{CIFNet} & \textbf{109.9 MB} & \textbf{109.9 MB} & $\mathcal{O}(C d^2)$ \\
    \bottomrule
\end{tabular}
}
\end{table}

The dominant cost for both ACIL and DS-AL is the covariance matrix, which accounts for over $90\%$ of their total element count in all configurations (Table~\ref{tab:memory_breakdown}). This term grows quadratically with $B$, which must typically be set to large values to achieve competitive classification accuracy.  As shown in Table~\ref{tab:memory_scaling}, CIFNet eliminates this dependency entirely: its classifier state is determined solely by the backbone dimension $d$ and the number of classes $C$, both of which are fixed prior to incremental deployment and do not grow with the task horizon or projection hyperparameters.

\end{document}